\documentclass[twoside,11pt]{article}

\usepackage{jmlr2e_mod}

\usepackage{amsmath}
\usepackage{amssymb}
\makeatletter
 \newcommand{\superimpose}[2]{{\ooalign{$#1\@firstoftwo#2$\cr\hfil$#1\@secondoftwo#2$\hfil\cr}}}
\makeatother

\newcommand{\cind}{\perp\hspace*{-1.35ex}\perp}
\newcommand{\ncind}{\not\cind}

\usepackage{graphicx}
\usepackage[linesnumbered]{algorithm2e}

\jmlrheading{1}{2014}{1-48}{4/00}{10/00}{Ricardo Silva and Robin Evans}

\ShortHeadings{Causal Inference through a Witness Protection Program}{Silva and Evans}
\firstpageno{1}

\begin{document}

\title{Causal Inference through a Witness Protection Program}

\author{\name Ricardo Silva \email ricardo@stats.ucl.ac.uk \\
       \addr Department of Statistical Science and CSML\\
       University College London\\
       London WC1E 6BT, UK
       \AND
       \name Robin Evans \email evans@stats.ox.ac.uk \\
       \addr Department of Statistics\\
       University of Oxford\\
       Oxford OX1 3TG, UK}

\editor{TBA}

\maketitle

\begin{abstract}
  One of the most fundamental problems in causal inference is the
  estimation of a causal effect when variables are confounded. This is
  difficult in an observational study,
  because one has no direct evidence that all confounders have
  been adjusted for. We introduce a novel approach for estimating
  causal effects that exploits observational conditional
  independencies to suggest ``weak'' paths in a unknown causal
  graph. The widely used faithfulness condition of Spirtes et al.\ is
  relaxed to allow for varying degrees of ``path cancellations'' that
  imply conditional independencies but do not rule out the
  existence of confounding causal paths. The outcome is a posterior
  distribution over bounds on the average causal effect via a linear
  programming approach and Bayesian inference. We claim this approach
  should be used in regular practice along with other default tools in
  observational studies.
\end{abstract}


\section{Contribution}

We provide a new methodology for obtaining bounds on the
average causal effect (ACE) of a variable $X$ on a variable $Y$.
For binary variables, the ACE is defined as
\begin{equation}
\label{eq:ace_first}
E[Y\, |\, do(X = 1)] - E[Y\, |\, do(X = 0)] = P(Y = 1\, |\, do(X = 1)) - P(Y = 1\, |\, do(X = 0)),
\end{equation}
where $do(\cdot)$ is the operator of \cite{pearl:00}, denoting
distributions where a set of variables has been intervened on by an
external agent. In this paper, we assume the reader is familiar with
the concept of causal graphs, the basics of the $do$ operator, and the
basics of causal discovery algorithms such as the PC algorithm of
\cite{sgs:00}. We provide a short summary for context in Section
\ref{sec:background}.

The ACE is in general not identifiable from observational data.  We
obtain upper and lower bounds on the ACE by exploiting a set of
(binary) covariates, which we also assume are not affected by $X$ or
$Y$ (justified by temporal ordering or other background assumptions).
Such covariate sets are often found in real-world problems, and
form the basis of many of the observational studies done in practice
\citep{rose:02}. However, it is not obvious how to obtain the
ACE as a function of the covariates. Our contribution modifies the
results of \cite{entner:13}, who exploit conditional independence
constraints to obtain point estimates of the ACE, but relying on
assumptions that might be unstable with finite sample sizes. Our
modification provides a different interpretation of their search
procedure, which we use to generate candidate {\it instrumental
variables}
\citep{manski:07}. The linear programming approach of \cite{dawid:03},
inspired by \cite{balke:97} and further refined by
\cite{ramsahai:12}, is then modified to generate bounds on the ACE
by introducing constraints on some causal paths, motivated as
relaxations of \cite{entner:13}. The new setup can be computationally
expensive, so we introduce further relaxations to the linear program
to generate novel symbolic bounds, and a fast algorithm that sidesteps
the full linear programming optimization with some simple, message
passing-like steps.

In Section \ref{sec:background}, we briefly discuss the background of
the problem. Section \ref{sec:wpp} contains our main methodology,
while commenting on why the unidentifiability of the ACE matters even
in a Bayesian context. Section \ref{sec:scale-up} discusses an
analytical approximation of the main results of the methodology, as
well as a way by which this provides scaling-up possibilities for the
approach. Our approach introduces free parameters, and Section
\ref{sec:parameters} provides practical guidelines on how to choose
them. Section
\ref{sec:experiments} contains experiments with synthetic and real
data.

\begin{figure}[t]
\begin{center}
\begin{tabular}{ccccc}
\includegraphics[width=1in]{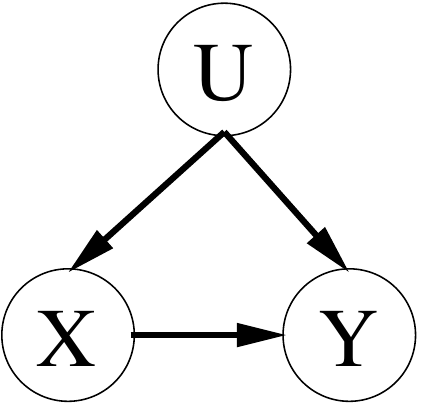} &
\includegraphics[width=1in]{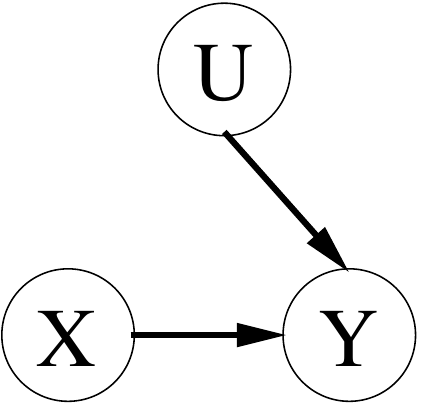} &
\includegraphics[width=1in]{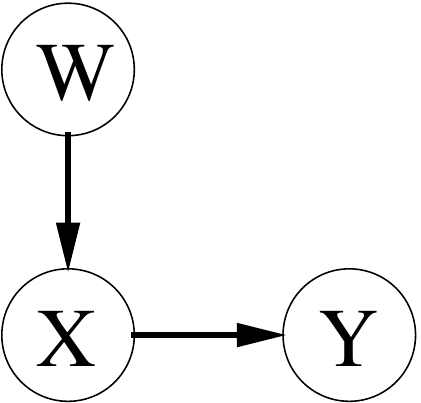} &
\includegraphics[width=1in]{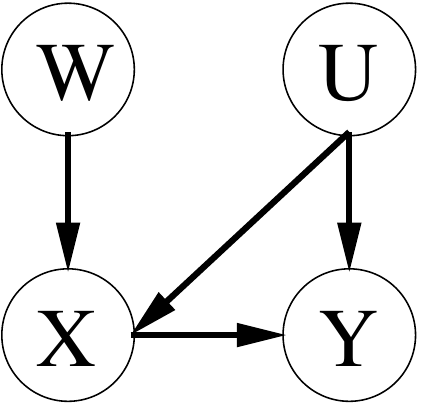} &
\includegraphics[width=1in]{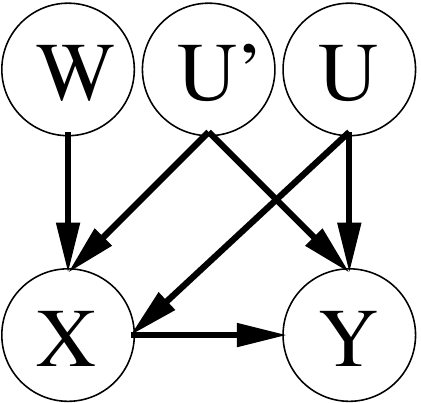}\\
(a) & (b) & (c) & (d) & (e)\\
\end{tabular}
\end{center}
\caption{(a) A generic causal graph where $X$ and $Y$ are confounded by some $U$.
(b) The same system in (a) where $X$ is intervened upon by an external agent.
(c) A system where $W$ and $Y$ are independent given $X$. 
(d) A system where it is possible to use faithfulness to discover that 
$U$ is sufficient to block all back-door paths between $X$ and $Y$. 
(e) Here, $U$ itself is not sufficient.}
\label{fig:summary}
\end{figure}

\section{Background: Instrumental Variables, Witnesses and Admissible Sets}
\label{sec:background}

Assuming $X$ is a potential cause of $Y$, but not the opposite, a
cartoon of the possibly complex real-world causal system containing
$X$ and $Y$ is shown in Figure \ref{fig:summary}(a). $U$ represents
the universe of common causes of $X$ and $Y$. In control and
policy-making problems, we would like to know what happens to the
system when the distribution of $X$ is overridden by some external
agent (e.g., a doctor, a robot or an economist). The resulting
modified system is depicted in Figure \ref{fig:summary}(b), and
represents the family of distributions indexed by $do(X = x)$: the
graph in (a) has undergone a ``surgery'' that removes incoming edges to $X$.
\cite{sgs:00} provide an account of the first graphical methods applying this idea, which are related to the overriding of structural equations
proposed by \cite{haavelmo:43}. Notice that if $U$ is observed in the dataset, then we
can obtain the distribution $P(Y = y\, |\, do(X = x))$ by simply
calculating $\sum_uP(Y = y\, |\, X = x, U = u)P(U = u)$
\citep{sgs:00}.  This was popularized by
\cite{pearl:00} as {\it back-door adjustment}. In general 
$P(Y = y\, |\, do(X = x))$ can be 
vastly different from $P(Y = y\, |\, X = x)$.

The ACE can usually be estimated via a trial in which $X$ is randomized: this is
equivalent to estimating the conditional distribution of $Y$ given $X$
under data generated as in Figure \ref{fig:summary}(b). In contrast, in an
{\it observational study} \citep{rose:02} we obtain data generated by
the system in Figure \ref{fig:summary}(a). If one believes all
relevant confounders $U$ have been recorded in the data then back-door
adjustment can be used, though such completeness is uncommon. 
By postulating knowledge of the causal graph relating
components of $U$, one can infer whether a measured subset of the
causes of $X$ and $Y$ is enough
\citep{pearl:00,ilya:11,pearl:09a}. Without knowledge of the causal
graph, assumptions such as {\it faithfulness} \citep{sgs:00}
are used to infer it.

The faithfulness assumption states that a conditional independence
constraint in the observed distribution exists if and only if a
corresponding structural independence exists in the underlying causal
graph. For instance, observing the independence $W \cind Y\, |\, X$,
and assuming faithfulness and the causal order, we can infer the
causal graph Figure \ref{fig:summary}(c); in all the other graphs this
conditional independence in not implied.
We deduce that no
unmeasured confounders between $X$ and $Y$ exist. This simple procedure
for identifying chains $W \rightarrow X \rightarrow Y$ is useful
in exploratory data analysis \citep{storey:07,cooper:97}, where a large number of 
possible causal relations $X \rightarrow Y$ are unquantified but can be screened
using observational data before experiments are performed. The purpose
of using faithfulness is to be able to identify such quantities.

\cite{entner:13} generalize the discovery of chain
models to situations where a non-empty set of covariates is necessary to block
all back-doors. Suppose $\mathcal W$ is a set of covariates
which are known not to be effects of either $X$ or $Y$, and we want
to find an {\it admissible set} contained in $\mathcal W$: a set of
observed variables which we can use for back-door adjustment to obtain
$P(Y = y\, |\, do(X = x))$. Entner et al.'s ``Rule 1'' states the following:\\

\noindent {\it {\bf Rule 1}: If there exists a variable
$W \in \mathcal W$ and a set $\mathbf Z \subseteq \mathcal W \backslash \{W\}$
such that:
\begin{align*}
\text{(i)} \quad & W \ncind Y\, |\, \mathbf Z 
&\text{(ii)} \quad & W \cind Y\, |\, \mathbf Z \cup \{X\}.
\end{align*}
\noindent then infer that $\mathbf Z$ is an admissible
set}\footnote{\cite{entner:13} aims also at identifying
zero effects with a ``Rule 2''. For simplicity of presentation, we
assume that the effect of interest was already identified as
non-zero.}.\\

A point estimate of the ACE can then be found using $\mathbf Z$. Given
that $(W, \mathbf Z)$ satisfies Rule 1, we call $W$ a {\it
  witness} for the admissible set $\mathbf Z$. The model in Figure
\ref{fig:summary}(c) can be identified with Rule 1, where $W$ is the
witness and $\mathbf Z = \emptyset$. In this case, a so-called
Na\"ive Estimator\footnote{Sometimes we use the word ``estimator'' to mean
a functional of the probability distribution instead of a statistical
estimator that is a function of samples of this distribution. Context should make it clear
when we refer to an actual statistic or a functional.} $P(Y = 1\, |\, X = 1) - P(Y = 1\, |\, X = 0$) will provide
the correct ACE. If $U$ is observable in Figure
\ref{fig:summary}(d), then it can be identified as an admissible set
for witness $W$. Notice that in Figure \ref{fig:summary}(a), taking
$U$ as a scalar, it is not possible to find a witness since there are
no remaining variables. Also, if in Figure \ref{fig:summary}(e) our
covariate set $\mathcal W$ is $\{W, U\}$, then no witness can be found since
$U'$ cannot be blocked. Hence, it is possible for a procedure based on
Rule 1 to answer ``I don't know'' even when a
back-door adjustment would be possible {\it if} one knew the causal
graph. However, using the faithfulness assumption alone one cannot 
do better: Rule 1 is complete for non-zero effects without
more information \citep{entner:13}.

Despite its appeal, the faithfulness assumption is not without
difficulties. Even if unfaithful distributions can be ruled out as
pathological under seemingly reasonable conditions \citep{meek:95},
distributions which lie close to (but not on) a simpler model may in
practice be indistinguishable from distributions within that simpler
model at finite sample sizes.  

To appreciate these complications, consider the structure in Figure
\ref{fig:summary}(d) with $U$ unobservable. 
Here $W$ is randomized but $X$ is not, 
and we would like to know the ACE of $X$ on $Y$\footnote{A
  classical example is in non-compliance: suppose $W$ is the assignment 
  of a patient to either
  drug or placebo, $X$ is whether the patient
  actually took the medicine or not, and $Y$ is a measure of
  health status. The doctor controls $W$ but not $X$. This 
  problem is discussed by
  \cite{pearl:00} and \cite{dawid:03}.}.  $W$ is 
sometimes known as an {\it instrumental variable} (IV), and  we call 
Figure \ref{fig:summary}(d) the {\it standard IV
  structure} (SIV): the distinctive features here being the constraints
$W \cind U$ and $W \cind Y\ |\ \{X, U\}$, statements which include latent
variables. If this structure is known, optimal bounds
\begin{equation*}
\mathcal L_{SIV} \leq E[Y\ |\ do(X = 1)] - E[Y\ |\ do(X = 0)] \leq \mathcal U_{SIV}
\end{equation*}

\noindent can be obtained without further assumptions, and estimated
using only observational data over the binary variables $W$, $X$ and
$Y$ \citep{balke:97}. However, there exist distributions faithful to the IV
structure but which at finite sample sizes may appear to satisfy the
Markov property for the structure $W \rightarrow X \rightarrow Y$; in
practice this can occur at any finite sample size \citep{robins:03}.
The true average causal effect may lie anywhere in the interval
$[\mathcal L_{SIV}, \mathcal U_{SIV}]$, which can be rather wide even when
$W \cind Y\, |\, X$, as shown by the following result:

\begin{proposition}
\label{prop:siv}
If $W \cind Y\, |\, X$ and the model follows the causal
structure of the standard IV graph, then 
$\mathcal U_{SIV} -  \mathcal L_{SIV} = 1 - |P(X = 1\ | W = 1) - P(X = 1\, |\, W = 0)|$.
\end{proposition}

\noindent All proofs in this manuscript are given in Appendix A.  For
a fixed joint distribution $P(W, X, Y)$, the length of such an interval
cannot be further improved \citep{balke:97}. Notice that the length of
the interval will depend on how strongly associated $W$ and $X$ are:
$W = X$ implies $\mathcal U_{IV} -  \mathcal L_{IV} = 0$ as expected,
since this is the scenario of a perfect intervention. The scenario where
$W \cind X$ is analogous to not having any instrumental variable,
and the length of corresponding interval is 1.

Thus, the true ACE may differ considerably from the Na\"ive Estimator,
appropriate for the simpler structure $W \rightarrow X \rightarrow Y$ but not
for the standard IV structure. While we emphasize that this is a
`worst-case scenario' analysis and by itself should not rule out
faithfulness as a useful assumption, it is desirable to provide a
method that gives greater control over violations of faithfulness.

\section{Methodology: the Witness Protection Program}
\label{sec:wpp}

The core of our idea is (i) to {\it invert the usage of Entner et al.'s Rule 1}, so that pairs
$(W, \mathbf Z)$ should provide an instrumental variable bounding method
instead of a back-door adjustment; (ii) express violations of faithfulness
as {\it bounded violations of local independence}; (iii) find bounds on
the ACE using {\it a linear programming formulation}.

Let $(W, \mathbf Z)$ be any pair found by a search procedure that
decides when Rule 1 holds. $W$ will play the role of an instrumental
variable, instead of being discarded. Conditional on $\mathbf Z$, the
lack of an edge $W \rightarrow Y$ can be justified by faithfulness (as
$W \cind Y\, |\, \{X, \mathbf Z\}$). For the same reason, there should
not be any (conditional) dependence between $W$ and a possible
unmeasured common parent\footnote{In this manuscript, we will sometimes
refer to $U$ as a {\it set} of common parents, although we do not
change our notation to bold face to reflect that.} $U$ of $X$ and
$Y$. Hence, $W \cind U$ and $W
\cind Y \ |\ \{U, X\}$ hold given $\mathbf Z$. A standard IV bounding procedure such as
\citep{balke:97} can then be used conditional on each individual value
$\mathbf z$ of $\mathbf Z$, then averaged over $P(\mathbf Z)$. That
is, we can independently obtain lower and upper bounds $\{\mathcal
L(\mathbf z),
\mathcal U(\mathbf z)\}$ for each value $\mathbf z$, and bound the ACE by
\begin{equation}
\label{eq:use_z}
\sum_{\mathbf z}\mathcal L(\mathbf z)P(\mathbf Z = \mathbf z) \leq 
E[Y\ |\ do(X = 1)] - E[Y\ |\ do(X = 0)] 
\leq \sum_{\mathbf z}\mathcal U(\mathbf z)P(\mathbf Z = \mathbf z),
\end{equation}
\noindent since $E[Y\ |\ do(X = 1)] - E[Y\ |\ do(X = 0)] 
= \sum_{\mathbf z} (E[Y\ |\ do(X = 1), \mathbf Z = \mathbf z] - E[Y\ |\ do(X = 0), \mathbf Z = \mathbf z])P(\mathbf Z = \mathbf z)$.

Under the assumption of faithfulness and the satisfiability of Rule 1,
the above interval estimator is redundant, as Rule 1 allows the direct use of the
back-door adjustment using $\mathbf Z$. Our goal is to not enforce faithfulness,
but use Rule 1 as a motivation to exclude arbitrary violations of faithfulness.

In what follows, assume $\mathbf Z$ is set to a particular value $\mathbf z$ and
all references to distributions are implicitly assumed to be defined conditioned on
the event $\mathbf Z = \mathbf z$. That is, for simplicity of
notation, we will neither represent nor condition on $\mathbf Z$ explicitly.
The causal ordering where $X$ and $Y$ cannot precede any other variable is also
assumed, as well as the causal ordering between $X$ and $Y$.

Consider a standard parameterization of a directed acyclic graph (DAG)
model, not necessarily causal, in terms of conditional probability tables (CPTs): let
$\theta_{v.\mathbf p}^V$ represent $P(V = v\ |\ Par(V) = \mathbf p)$
where $V \in \{W, X, Y, U\}$ denotes both a random variable and a
vertex in the corresponding DAG; $Par(V)$ is the corresponding
set of parents of $V$. Faithfulness violations occur when independence
constraints among observables are not {\it structural}, but due to
``path cancellations.'' This means that parameter values are arranged so 
that $W \cind Y\ |\ X$ holds, but paths connecting
$W$ and $U$, or $W$ and $Y$, may exist so that 
either $W \not\cind U$ or $W \not\cind Y \ |\ \{U, X\}$.
In this situation, some combination of the following
should hold true:
\begin{equation}
\begin{array}{rcl}
P(Y = y\ |\ X = x, W = w, U = u) & \neq & P(Y = y\ |\ X = x, U = u)\\
P(Y = y\ |\ X = x, W = w, U = u) & \neq & P(Y = y\ |\ X = x, W = w)\\
P(X = x\ |\ W = w, U = u)        & \neq & P(X = x\ |\ W = w)\\
P(U = u\ |\ W = w)               & \neq & P(U = u),\\
\end{array}
\label{eq:basic_viol}
\end{equation}
\noindent for some $\{w, x, y, u\}$ in the sample space of $P(W, X, Y, U)$.

For instance, if the second and third statements above are true and under the
assumption of faithfulness, this implies the existence of
an active path into $X$ and $Y$ via $U$, conditional on
$W$\footnote{That is, a path that d-connects $X$ and $Y$ and includes
$U$, conditional on $W$; it is ``into'' $X$ (and $Y$) because the edge
linking $X$ to the path points to $X$. See \cite{sgs:00} and
\cite{pearl:00} for formal definitions and more examples.}, such as $X
\leftarrow U \rightarrow Y$.  If the first statement is true, this
corresponds to an active path between $W$ and $Y$ into $Y$ that is not
blocked by $\{X, U\}$. If the fourth statement is true, $U$ and $W$
are marginally dependent, with a corresponding active path. Notice
some combinations are still compatible with a model where $W \cind
U$ and $W \cind Y \ |\ \{U, X\}$ hold: if the second statement in
(\ref{eq:basic_viol}) is false, this means $U$ cannot be a common
parent of $X$ and $Y$. This family of models is observationally
equivalent\footnote{Meaning a family of models where $P(W, X, Y)$ satisfies
the same constraints.} to one where $U$ is independent of all
variables.

When translating the conditions (\ref{eq:basic_viol}) into
parameters $\{\theta_{v.\mathbf p}^V\}$, we need to define which
parents each vertex has. In our CPT factorization, we define $Par(X)
= \{W, U\}$ and $Par(Y) = \{W, X, U\}$; the joint distribution of
$\{W, U\}$ can be factorized arbitrarily. In the next subsection, we
refine the parameterization of our model by introducing redundancies:
we provide a parameterization for the latent variable model $P(W, X,
Y, U)$, the interventional distribution $P(W, Y, U\ |\ do(X))$ and the
corresponding (latent-free) marginals $P(W, X, Y)$, $P(W, Y\ |\
do(X))$. These distributions parameters are related, and cannot 
differ arbitrarily. It is this fact
that will allow us to bound the ACE
using only $P(W, X, Y)$.

\begin{figure}
\begin{center}
\includegraphics[width=2.5in]{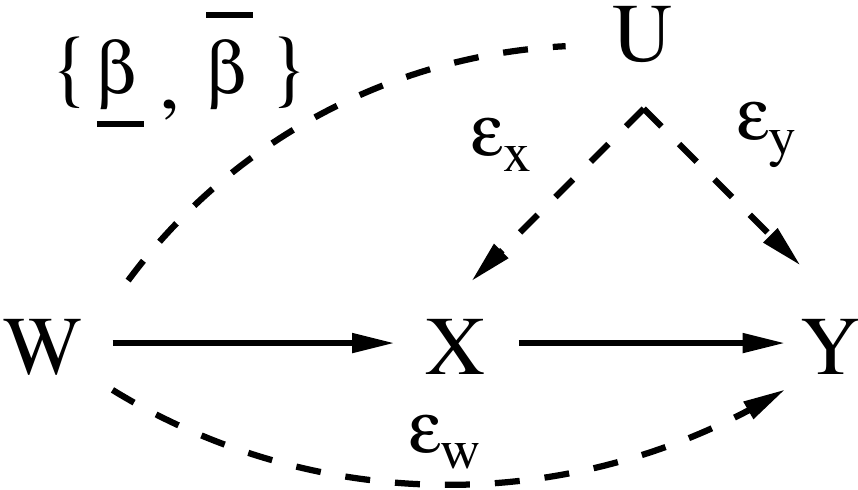}
\end{center}
\caption{A visual depiction of the family of assumptions introduced in our framework. Dashed
edges correspond to conditional dependencies that are constrained according to free parameters,
displayed along each corresponding edge. This is motivated by observing $W \cind Y\ |\ X$.}
\label{fig:w_relations}
\end{figure}

\subsection{Encoding Faithfulness Relaxations with Linear Constraints}
\label{sec:encoding}

We define a {\it relaxation of faithfulness} as any set of assumptions
that allows the relations in (\ref{eq:basic_viol}) to be true, but not
necessarily in an {\it arbitrary} way: this means that while the
left-hand and right-hand sides of each entry of (\ref{eq:basic_viol})
are indeed different, their difference is bounded by either the
absolute difference or by ratios. Without such restrictions,
(\ref{eq:basic_viol}) will only imply vacuous bounds of length 1, as
discussed in our presentation of Proposition \ref{prop:siv}.

Consider the following parameterization of the distribution of $\{W,
X, Y, U\}$ under the observational and interventional regimes, and
their respective marginals obtained by integrating $U$ way\footnote{
Notice from the development in this Section that $U$ is not necessarily
a scalar, nor discrete.}. Again
we condition everywhere on a particular value $\mathbf
z$ of $\mathbf Z$ but, for simplicity of presentation, 
we supress this from our notation, since it is
not crucial to developments in this Section:
\vspace{-0.3in}
\begin{center}
\[
\begin{array}{rcl}
\zeta_{yx.w}^\star & \equiv& P(Y = y, X = x \, |\, W = w, U)\\
\zeta_{yx.w} & \equiv& \sum_U P(Y = y, X = x \, |\, W = w, U)P(U\, |\, W = w)\\
           & = & P(Y = y, X = x\, |\, W = w)\\[-6pt]
\\
\eta_{xw}^\star & \equiv& P(Y = 1\, |\, X = x, W = w, U)\\
\eta_{xw} & \equiv& \sum_U P(Y = 1\, |\, X = x, W = w, U)P(U\, |\, W = w)\\
         & = & P(Y = 1\, |\, do(X = x), W = w)\\[-6pt]
\\
\delta_{w}^\star & \equiv& P(X = 1\, |\,  W = w, U)\\
\delta_{w} & \equiv& \sum_U P(X = x \, |\,  W = w, U)P(U\, |\, W = w)\\ 
&=& P(X = 1\, |\,  W = w).\\
\end{array}
\]
\end{center}

\noindent Under this encoding, the ACE is given by 
\begin{equation}
\label{eq:ace}
\eta_{11}P(W = 1) + \eta_{10}P(W = 0) - \eta_{01}P(W = 1) - \eta_{00}P(W = 0).
\end{equation}
\noindent Notice that we do not explicitly parameterize the marginal of $U$, for
reasons that will become clear later.

We introduce the following assumptions, as illustrated by Figure \ref{fig:w_relations}:
\begin{align}
\label{eq:YW_bound}
|\eta_{x1}^{\star} - \eta_{x0}^{\star}| &\leq \epsilon_w \\
\label{eq:eps_y}
|\eta_{xw}^{\star}  - P(Y = 1\, |\, X = x, W = w)| &\leq \epsilon_y \\
\label{eq:eps_x}
|\delta_{w}^{\star}  - P(X = 1\, |\, W = w)| &\leq \epsilon_x\\
\label{eq:UW_bound2}
\underline{\beta}P(U) \leq P(U\, |\, W = w) &\leq \bar{\beta}P(U).
\end{align}
%
\noindent Setting $\epsilon_w = 0$, $\underline \beta = \bar \beta =
1$ recovers the standard IV structure. Further assuming $\epsilon_y =
\epsilon_x = 0$ recovers the chain structure $W \rightarrow X
\rightarrow Y$. Under this parameterization in the case $\epsilon_y
= \epsilon_x = 1$, $\underline \beta = \bar \beta = 1$, 
\cite{ramsahai:12}, extending \cite{dawid:03}, used
linear programming to obtain bounds on the ACE. We will briefly describe
the four main steps of the framework of \cite{dawid:03}, and refer to the cited papers for more
details of their implementation. 

For now, assume that $\zeta_{yx.w}$ and $P(W = w)$ are known constants---that 
is, treat $P(W, X, Y)$ as known. This assumption will be
   dropped later. Dawid's formulation of a
bounding procedure for the ACE is as follows.\\

\noindent {\bf Step 1} {\it Notice that parameters $\{\eta_{xw}^\star\}$
take values in a 4-dimensional polytope.
Find the extreme points of this polytope. Do the same for $\{\delta_w^\star\}$.}\\

In particular, for $\epsilon_w = \epsilon_y = 1$, the polytope of
feasible values for the four dimensional vector $(\eta_{00}^\star,
\eta_{01}^\star, \eta_{10}^\star, \eta_{11}^\star)$ is the unit
hypercube $[0, 1]^4$, a polytope with a total of 16 vertices $(0, 0, 0,
0), (0, 0, 0, 1), \dots (1, 1, 1, 1)$. \cite{dawid:03} covered the
case $\epsilon_w = 0$, where a two-dimensional vector
$\{\eta_x^\star\}$ replaces $\{\eta_{xw}^\star\}$. In
\cite{ramsahai:12}, the case $0 \leq \epsilon_w < 1$ is also covered: some
of the corners in $[0, 1]^4$ disappear and are replaced by others. The case
where $\epsilon_w = \epsilon_x = \epsilon_y = 1$ is vacuous, in the sense that
the consecutive steps cannot infer non-trivial constraints on the ACE.\\

\noindent {\bf Step 2} {\it 
Find the extreme points of the joint space $\{\zeta_{yx.w}^\star\} \times \{\eta_{xw}^\star\}$ by mapping them
from the extreme points of $\{\delta_w^\star\} \times \{\eta_{xw}^\star\}$, since 
$\zeta_{yx.w}^\star = (\delta_w^\star)^x(1 - \delta_w^\star)^{(1 - x)}\eta_{xw}^\star$.}\\

The extreme points of the joint space $\{\delta_w^\star\} \times
\{\eta_{xw}^\star\}$ are just the combination of the extreme points of
each space. Some combinations $\delta_x^\star \times
\eta_{xw}^\star$ map to the same $\zeta_{yx.w}^\star$, while the
mapping from a given $\delta_x^\star \times \eta_{xw}^\star$ to
$\eta_{xw}^\star$ is just the trivial projection. At this stage, we obtain all the extreme
points of the polytope $\{\zeta_{yx.w}^\star\} \times \{\eta_{xw}^\star\}$ that are entailed
by the factorization of $P(W, X, Y, U)$ and our constraints. \\

\noindent {\bf Step 3} {\it Using the extreme points of the joint space
  $\{\zeta_{yx.w}^\star\} \times \{\eta_{xw}^\star\}$, find the
  dual polytope of this space in terms of linear inequalities.
  Points in this polytope are convex combinations of 
  $\{\zeta_{yx.w}^\star\} \times \{\eta_{xw}^\star\}$, shown by \cite{dawid:03} to correspond
  to the marginalizations over arbitrary $P(U)$. This results in constraints over
  $\{\zeta_{yx.w}\} \times \{\eta_{xw}\}$.}\\

This is the core step in \cite{dawid:03}: points in the polytope 
$\{\zeta_{yx.w}^\star\} \times \{\eta_{xw}^\star\}$ correspond to different marginalizations of $U$
according to different $P(U)$. Describing the polytope in terms of inequalities provides
all feasible distributions that result from marginalizing $U$ according to some $P(U)$. Because 
we included both $\zeta_{yx.w}^\star$ and $\eta_{xw}^\star$ in the same space, this will tie together $P(Y, X\ | W)$ and
$P(Y\ |\ do(X), W)$. \\

\noindent {\bf Step 4} {\it Finally, maximize/minimize (\ref{eq:ace}) with respect to $\{\eta_{xw}\}$ subject
  to the constraints found in Step 3 to obtain upper/lower bounds on the ACE.}\\

Allowing for the case where $\epsilon_x < 1$ or $\epsilon_y < 1$ is
just a matter of changing the first step, where box constraints are
set on each individual parameter as a function of the known $P(Y = y,
X = x\, |\, W = w)$, prior to the mapping in Step 2. The resulting
constraints are now implicitly non-linear in $P(Y = y, X = x\, |\, W =
w)$, but at this stage this does not matter as the distribution of the
observables is treated as a constant. That is, each resulting
constraint in Step 3 is a linear function of $\{\eta_{xw}\}$ and a
multilinear function on $\{\{\zeta_{yx.w}\}, \epsilon_x, \epsilon_y,
\epsilon_w, \bar\beta,
\underline\beta, P(W)\}$, as discussed in Section \ref{sec:scale-up}. 
Within the objective function (\ref{eq:ace}), the only decision
variables are $\{\eta_{xw}\}$, and hence Step 4 still sets up a linear
programming problem even if there are multiplicative interactions
between $\{\zeta_{yx.w}\}$ and parameters of constraints.

To allow for the case $\underline \beta < 1 < \bar \beta$, 
we substitute every occurrence of
$\zeta_{yx.w}$ in the constraints by $\kappa_{yx.w}
\equiv \sum_U \zeta_{yx.w}^\star P(U)$; notice
the difference between $\kappa_{yx.w}$ and $\zeta_{yx.w}$. Likewise,
we substitute every occurrence of $\eta_{xw}$ in the constraints by 
$\omega_{xw} \equiv \sum_U \eta_{xw}^\star P(U)$. Instead of
plugging in constants for the values of $\kappa_{yx.w}$ and turning
the crank of a linear programming solver, we treat $\{\kappa_{yx.w}\}$
(and $\{\omega_{xw}\}$) as unknowns, linking them to observables and
$\eta_{xw}$ by the constraints 
\begin{equation}
\begin{array}{cc}
\kappa_{yx.w} \leq \zeta_{yx.w} / \underline \beta & \kappa_{yx.w} \geq \zeta_{yx.w} / \bar \beta \\
\omega_{xw} \leq \eta_{xw} / \underline \beta      & \omega_{xw} \geq \eta_{xw} / \bar \beta \\
\end{array}
\end{equation}
\begin{equation}
\sum_{yx} \kappa_{yx.w} = 1.
\end{equation}

Finally, the steps requiring finding extreme points and converting
between representations of a polytope can be easily implemented using
a package such as Polymake\footnote{http://www.poymake.org} or the
{\sc scdd} package\footnote{http://cran.r-project.org/} for
{\sc R}. Once bounds are obtained for each particular value of $\mathbf
Z$, Equation (\ref{eq:use_z}) is used to obtain the unconditional bounds
assuming $P(\mathbf Z)$ is known.

In Section \ref{sec:parameters}, we provide some guidance on how to choose the free
parameters of the relaxation. However, it is relevant to point out that any choice of $\epsilon_w \geq 0,
\epsilon_y \geq 0, \epsilon_x \geq 0, 0 \leq \underline \beta \leq 1
\leq \bar \beta$ is {\it guaranteed to provide bounds that are at
  least as conservative} as the back-door adjusted point estimator of
\cite{entner:13}, which is always covered by the bounds. Background
knowledge, after a user is suggested a witness and admissible set, can
also be used to set relaxation parameters.

\begin{algorithm}[t]
 \SetKwInOut{Input}{input}
 \SetKwInOut{Output}{output}
 \Input{Binary data matrix $\mathcal D$; 
set of relaxation parameters $\aleph$;
covariate index set $\mathcal W$; cause-effect indices $X$ and $Y$}
 \Output{A set of triplets $(W, \mathbf Z, \mathcal B)$, where $(W, \mathbf Z)$ is a witness-admissible set pair contained in $\mathcal W$ and 
         $\mathcal B$ is a distribution  over lower/upper bounds on the ACE implied by the pair}
 \BlankLine
 $\mathcal R \leftarrow \emptyset$\;
 \For{each $W \in \mathcal W$}{
  \For{every admissible set $\mathbf Z \subseteq \mathcal W \backslash \{W\}$ identified by $W$ and $\aleph$ given $\mathcal D$}{
    $\mathcal B \leftarrow$ posterior over lower/upper bounds on the ACE as given by $(W, \mathbf Z, X, Y, \mathcal D, \aleph)$\;
    \If{there is no evidence in $\mathcal B$ to falsify the $(W, \mathbf Z, \aleph)$ model}{
       $\mathcal R \leftarrow \mathcal R \cup \{(W, \mathbf Z, \mathcal B)\}$\;
    }
  }
 }
 \Return{$\mathcal R$}
 \BlankLine
\caption{The outline of the Witness Protection Program algorithm.}
\label{tab:algo}
\end{algorithm}

So far, the linear programming formulated through Steps 1--4 assumes
one has already identified an appropriate witness $W$ and admissible set $\mathbf Z$, and that
the joint distribution $P(W, X, Y, \mathbf Z)$ is known. In the next
Section, we discuss how this procedure is integrated with statistical
inference for $P(W, X, Y, \mathbf Z)$ and the search procedure of \cite{entner:13}.  As the
approach provides the witness a degree of protection against
faithfulness violations, using a linear program, we call this
framework the {\it Witness Protection Program} ({\sc WPP}).

\subsection{Bayesian Learning and Result Summarization}
\label{sec:bayes_learn}

In the previous section, we treated (the conditional) $\zeta_{yx.w}$ and $P(W = w)$ as known.
A common practice is to replace them by plug-in estimators (and in the
case of a non-empty admissible set $\mathbf Z$, an estimate of
$P(\mathbf Z)$ is also necessary). Such models can also be falsified,
as the constraints generated are typically only supported by a strict
subset of the probability simplex. In principle, one could fit
parameters without constraints, and test the model by a direct check
of satisfiability of the inequalities using the plug-in values. 
However, this does not take into account the uncertainty in the
estimation. For the standard IV model, \cite{ramsahai:11} discuss a
proper way of testing such models in a frequentist sense.

Our models can be considerably more complicated. Recall that
constraints will depend on the extreme points of the
$\{\zeta_{yx.w}^\star\}$ parameters.  As implied by (\ref{eq:eps_y})
and (\ref{eq:eps_x}), extreme points will be functions of
$\zeta_{yx.w}$. Writing the constraints fully in terms of the observed
distribution will reveal non-linear relationships.  We approach the
problem in a Bayesian way. We will assume first the dimensionality of
$\mathbf Z$ is modest (say, 10 or less), as this is the case in most
applications of faithfulness to causal discovery. We parameterize
$\zeta_{yxw}^{\mathbf z} \equiv P(Y = y, X = x, W = w\ |\ \mathbf Z = \mathbf z)$ as a full $2
\times 2 \times 2$ contingency table\footnote{That is, we allow for
  dependence between $W$ and $Y$ given $\{X, \mathbf Z\}$,
  interpreting the decision of independence used in Rule 1 as being
  only an indicator of approximate independence.}. In the context of
the linear programming problem of the previous Section,
for a given $\mathbf z$, we have $\zeta_{yx.w} = \zeta_{yxw} / P(W = w)$,
$P(W = w) = \sum_{yx}\zeta_{yxw}$.

Given that the dimensionality of the problem is modest, we assign to
each three-variate distribution $P(Y, X, W\, |\, \mathbf Z = \mathbf
z)$ an independent Dirichet prior for every possible assigment of
$\mathbf Z$, constrained by the inequalities implied by the
corresponding polytopes. The posterior is also a 8-dimensional
constrained Dirichlet distribution, where we use rejection sampling to
obtain a posterior sample by proposing from the unconstrained
Dirichlet. A Dirichlet prior is also assigned to $P(\mathbf Z)$.
Using a sample from the posterior of
$P(\mathbf Z)$ and a sample (for each possible value $\mathbf z$) from
the posterior of $P(Y, X, W\, |\, \mathbf Z = \mathbf z)$, we obtain a
sample upper and lower bound for the ACE by just running the linear
program for each sample of $\{\eta_{yxw}^\mathbf z\}$ and $\{P(\mathbf Z = \mathbf z)\}$.

The full algorithm is shown in Algorithm \ref{tab:algo}, where $\aleph
\equiv \{\epsilon_w, \epsilon_x, \epsilon_y, \underline \beta, \bar
\beta\}$. The search procedure is left unspecified, as different
existing approaches can be plugged into this step. See
\cite{entner:13} for a discussion. In Section \ref{sec:experiments} we
deal with small dimensional problems only, using the brute-force
approach of performing an exhaustive search for $\mathbf Z$.  In
practice, brute-force can be still valuable by using a method such as
discrete PCA \citep{buntine:04} to reduce $\mathcal W \backslash
\{W\}$ to a small set of binary variables.  To decide whether the
premises in Rule 1 hold, we merely perform Bayesian model selection
with the BDeu score
\citep{buntine:91} between the full graph $\{W \rightarrow X, W
\rightarrow Y, X \rightarrow Y\}$ (conditional on $\mathbf Z$) and the
graph with the edge $W \rightarrow Y$ removed. 

Step 5 in Algorithm \ref{tab:algo} is a ``falsification test.''  Since
the data might provide a bad fit to the constraints entailed by the
model, we opt not to accept every pair $(W, \mathbf Z)$ that passes Rule
1.  One possibility is to calculate the posterior distribution of the
model where constraints are enforced, and compare it against the
posteriors of the saturated model given by the unconstrained
contingency table.  This requires another prior over the constraint
hypothesis and the calculation of the corresponding marginal
likelihoods. As an alternative approach, we adopt the pragmatic rule
of thumb suggested by \cite{richardson:11}: sample $M$ samples from
the $\{\zeta_{yxw}^{\mathbf z}\}$ posterior given the {\it unconstrained} model,
and check the proportion of values that are rejected. If more than
$95\%$ of them are rejected, we take this as an indication that the
proposed model provides a bad fit and reject the given choice of $(W,
\mathbf Z)$.

The final result provides a set of posterior distributions over
bounds, possibly contradictory, which should be summarized as
appropriate. One possibility is to check for the union of all
intervals or, as a simpler alternative, report the lowest of the lower
bound estimates and the highest of the upper bound estimates using a
point estimate for each bound\footnote{One should not confuse credible
intervals with ACE intervals, as these are two separate concepts: each
lower or upper bound is a function of the unknown $P(W, X, Y, \mathbf
Z)$ and needs to be estimated. There is posterior uncertainty over
each lower/upper bound as in any problem where a functional of a
distribution needs to be estimated. So the posterior distribution and
the corresponding {\it credible intervals} over the {\it ACE
intervals} are perfectly well-defined as in any standard Bayesian
inference problem.}:
\begin{enumerate}
\item for each $(W, \mathbf Z)$ in $\mathcal R$, calculate the posterior expected
value of the lower and upper bounds;
\item report the interval $\mathcal L \leq ACE \leq \mathcal U$ where $\mathcal L$ is the
minimum of the lower bounds and $\mathcal U$ the maximum of the upper bounds.
\end{enumerate}
\noindent Alternatively to using the expected posterior estimator for the lower/upper bounds, one can, for instance, report the 0.025
quantile of the marginal lower bound distribution and the 0.975
quantile of the marginal upper bound distribution. Notice, however, this
does not give a 0.95 credible interval over ACE intervals
as the lower bound and the upper bound are dependent in the posterior.

In our experiments, we use a different summary. As we calculate the
log-marginal posterior $M_1, M_2, M_3, M_4$ for the hypotheses $W
\ncind Y\ |\ \mathbf Z$, $W \cind Y\ |\
\mathbf Z$, $W \cind Y\ |\ \mathbf Z \cup \{X\}$, $W \ncind
Y\ |\ \mathbf Z \cup \{X\}$, respectively, we
use the score 
\begin{equation}
\label{eq:wpp_score}
(M_1 - M_2) + (M_3 - M_4)
\end{equation}
\noindent to assess the quality of the
bounds obtained with the corresponding witness-admissible set pair.
We then report the corresponding interval and evaluation metric
based on this criterion.

\subsection{A Note on Weak Dependencies}
\label{sec:app_ind}

As we briefly mentioned in the previous Section, our parameterization
$\{\zeta_{yxw}^{\mathbf z}\}$ does not enforce the independence
condition $W \cind Y\ |\ \mathbf Z \cup \{X\}$ required by Rule 1. Our
general goal is to let {\sc WPP} accept ``near independencies,'' in
which the meaning of the symbol $\cind$ in practice means weak
dependence\footnote{The procedure that decides conditional
independencies in Section \ref{sec:bayes_learn} is a method for
testing exact independencies, although the prior on the independence
assumption regulates how strong the evidence in the data should be for
independence to be accepted.}.  We do not define what a weak
dependence should mean, except for the general guideline that some
agreed measure of conditional association should be ``small.'' Our
pragmatic view on {\sc WPP} is that Rule 1, when supported by weak
dependencies, should be used as a motivation for the constraints in
Section \ref{sec:encoding}. That is, the assumption that ``weak
dependencies are not generated by arbitrary near-path cancellations,''
reflecting the belief that very weak associations should correspond to
weak direct causal effects (and, where this is unacceptable, {\sc WPP}
should either be adapted to exclude relevant cases, or not be
used). At the same time, users of {\sc WPP} do not need to accept this
view, as the method does not change under the usual interpretation of
$\cind$, but computational gains can be obtained by using a
parameterization that encodes the independence.

\begin{figure}[t]
\begin{center}
\begin{tabular}{ccc}
\includegraphics[width=2in]{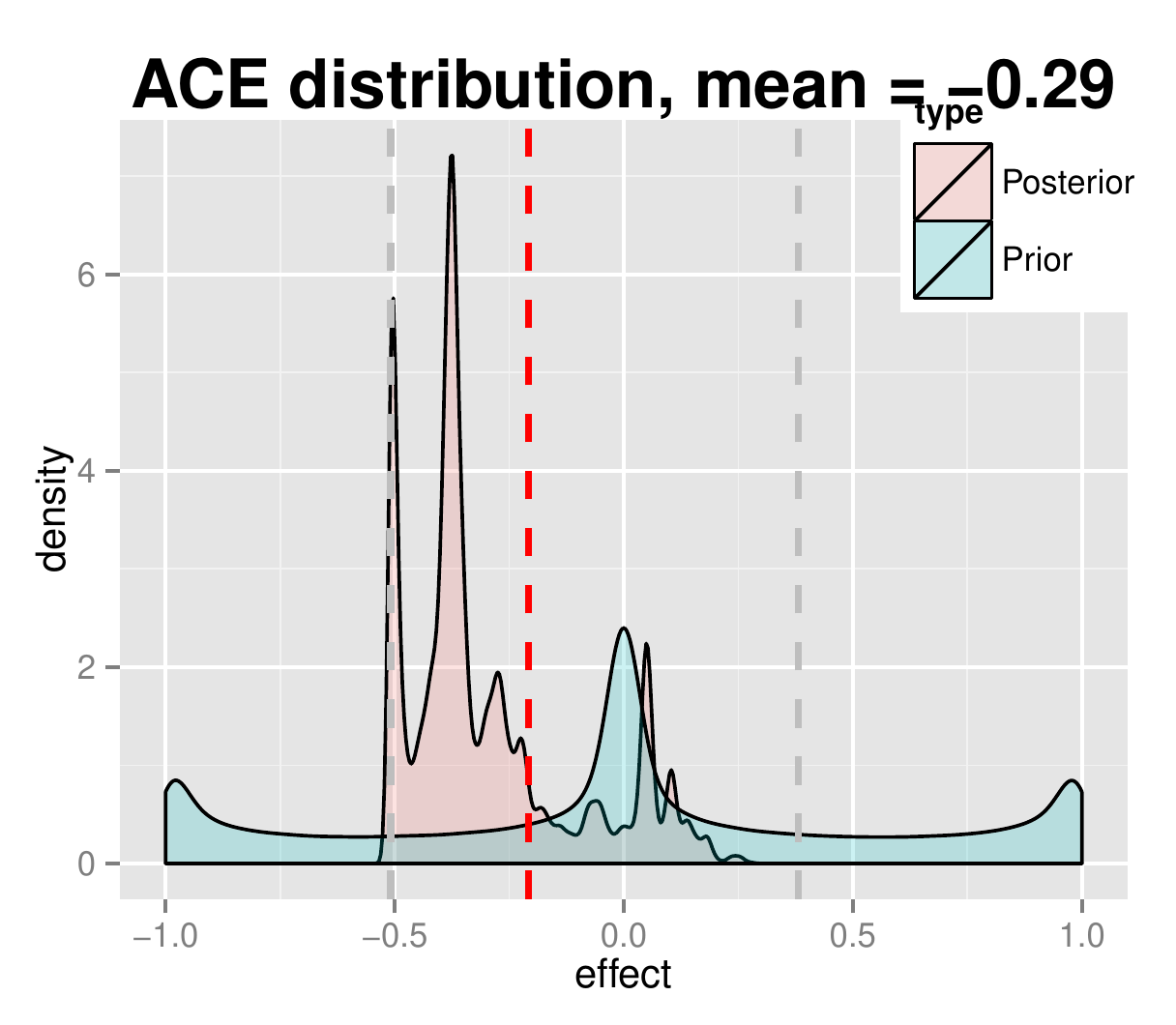} & \hspace{-0.2in}
\includegraphics[width=2in]{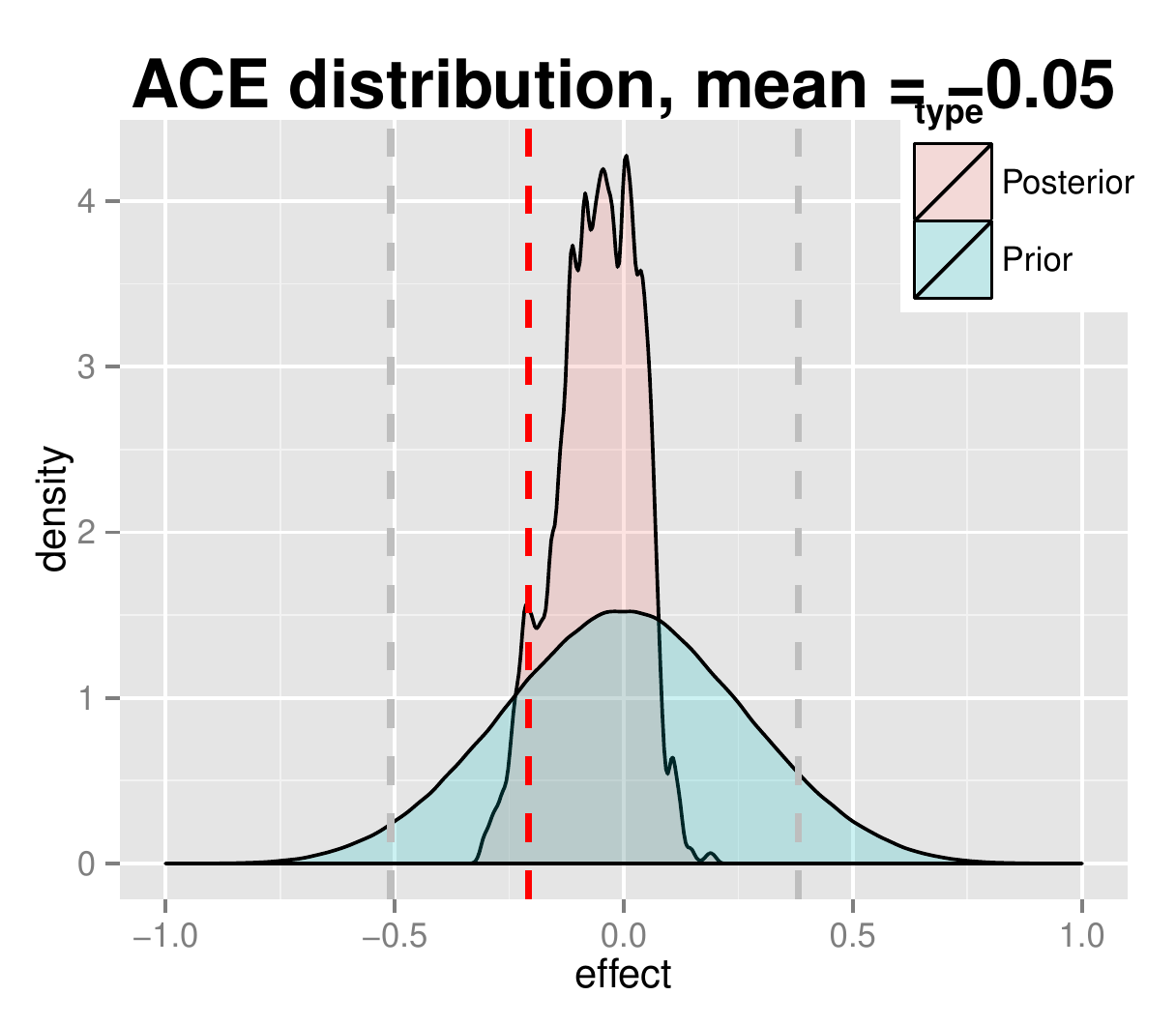} & \hspace{-0.2in}
\includegraphics[width=2in]{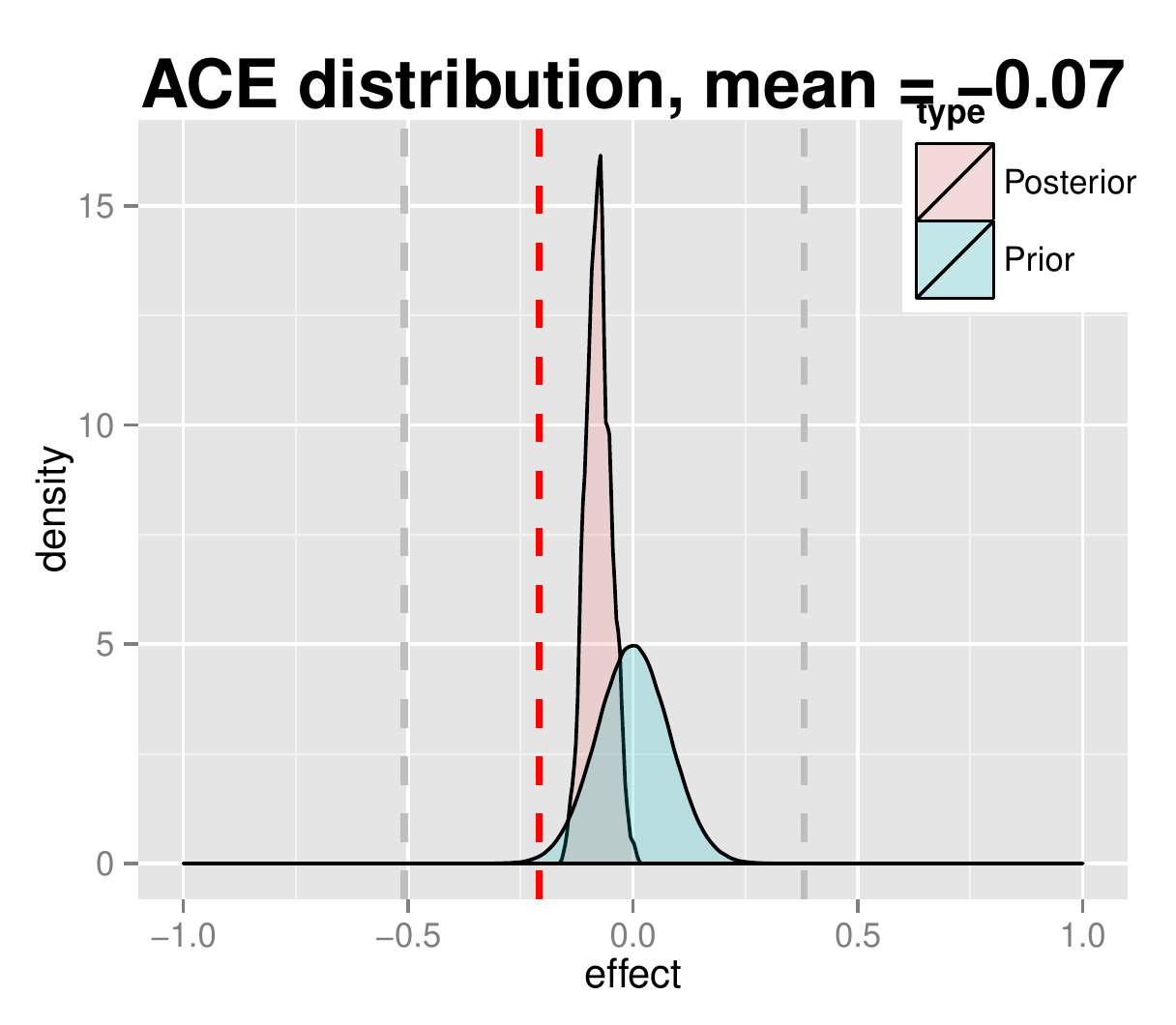} \\
(a) & (b) & (c)
\end{tabular}
\end{center}
\caption{Posterior over the ACE obtained by three different priors
conditioned on a synthetic dataset of size 1,000,000.
Posterior computed by running 1,000,000 iterations of Gibbs sampling.
The (independent) priors for $\theta_{1.xu}^Y$ and $\theta_{x.wu}^X$
are Beta $(\alpha, \alpha)$, while $\theta_u^U$ is given a Dirichlet
$(\alpha, \alpha, \alpha, \alpha)$. We set $\alpha = 0.1, 1, 10$ for
the cases shown in (a), (b) and (c), respectively. Vertical red line shows
the true ACE, while the population IV bounds are shown with grey lines. As the prior
gets less informative (moving from (c) to (a)), the erratic shape of the
posterior distribution also shows the effect of bad Gibbs sampling mixing.
Even with a very large dataset, the concentration of the posterior is highly
dependent on the concentration of the prior.}
\label{fig:lvm1}
\end{figure}

\subsection{A Note on Unidentifiability}
\label{sec:unidentifiability}

An alternative to bounding the ACE or using back-door adjustments is to
put priors directly on the latent variable model for $\{W, X, Y, U\}$.
Using the standard IV model as an example, we can define parameters
$\theta_{y.xu}^Y \equiv P(Y = y\ |\ X = x, U = u)$,
$\theta_{x.wu}^X \equiv P(X = x\ |\ W = w, U = u)$ and
$\theta_u^U \equiv P(U = u)$, on which priors are imposed\footnote{$P(W = w)$ is not necessary, as the
standard IV bounds \cite{balke:97} do not depend on it.}. No complicated
procedure for generating constraints in the observable marginal is necessary,
and the approach provides point estimates of the ACE instead of bounds.

This sounds too good to be true, and indeed it is: results
strongly depend on the prior, regardless of sample size. To illustrate
this, consider a simulation from a standard IV model (Figure
\ref{fig:summary}(c)), with $\mathbf Z =
\emptyset$ and $U$ an unobservable discrete variable of 4 levels. 
We generated a model by setting $P(W = w) =
0.5$ and sampling parameters $\theta_{1.xu}^Y$ and $\theta_{1.wu}^X$
from the uniform $[0, 1]$ distribution, while the 4-dimensional vector
$\theta_u^U$ comes from a Dirichlet $(1, 1, 1, 1)$. The resulting
model had an ACE of $-0.20$, with a wide IV interval $[-0.50, 0.38]$
as given by the method of \cite{balke:97}. Narrower intervals can only
be obtained by making more assumptions: there is no free
lunch. However, as in this case where {\sc WPP} cannot identify any
witness, one might put priors on the latent variable model to get a
point estimate, such as the posterior expected value of the ACE.

To illustrate the pitfalls of this approach, we perform
Bayesian inference by putting priors directly on the CPT parameters of
the latent variable model, assuming we know the correct number of
levels for $U$. Figure \ref{fig:lvm1} shows some results with a few
different choices of priors. The sample size is large enough so that
the posterior is essentially entirely within the population bounds and
the estimation of $P(W, X, Y, Z)$ is itself nearly exact. The
posterior over the ACE covers a much narrower area than the IV
interval, but its behaviour is erratic.

This is not to say that informative priors on a latent variable model
cannot produce important results. For instance, \cite{steen:04}
discuss how empirical priors on smoking habits among blue-collar
workers were used in their epidemiological question: the causal effect
of the occupational harzard of silica exposure on lung cancer
incidence among industrial sand workers. Smoking is a confounding
factor given the evidence that smoking and occupation are
associated. The issue was that smoking was unrecorded among the
workers, and so priors on the latent variable relationship to the
observables were necessary. Notice, however, that this informative
prior is essentially a way of performing a back-door adjustment when
the adjustment set $\mathbf Z$ and treatment-outcome pair $\{X, Y\}$
are not simultaneously measured within the same subjects. When latent
variables are ``unknown unknowns,'' a prior on $P(Y\ |\ X, U)$ may be
hard to justify. \cite{richardson:11} discuss more issues on priors
over latent variable models as a way of obtaining ACE point estimates,
one alternative being the separation of identifiable and
unindentifiable parameters to make transparent the effect of prior
(mis)specification.

\section{Algebraic Bounds and the Back-substitution Algorithm}
\label{sec:scale-up}

Posterior sampling is expensive within the context of Bayesian {\sc
  WPP}: constructing the dual polytope for possibly millions of
instantiations of the problem is time consuming, even if each problem
is small.  Moreover, the numerical procedure described in Section
\ref{sec:wpp} does not provide any insight on how the different free
parameters $\{\epsilon_w, \epsilon_x, \epsilon_y, \underline \beta,
\bar \beta\}$ interact to produce bounds, unlike the analytical
bounds available in the standard IV case.  
\cite{ramsahai:12} derives analytical bounds under 
(\ref{eq:YW_bound}) given a {\it fixed}, {\it numerical} value of
$\epsilon_w$. We know of no previous analytical bounds as an algebraic
function of $\epsilon_w$.

\subsection{Algebraic Bounds}
\label{sec:bounds}

We derive a set of bounds, whose validity are proved by three theorems.
The first theorem derives separate upper and lower bounds on
$\omega_{xw}$ using all the assumptions except Equation
(\ref{eq:YW_bound}); this means constraints which do not link
distributions under different values of $W = w$. The second theorem
derives linear constraints on $\{\omega_{xw}\}$ using
(\ref{eq:YW_bound}) and more elementary constraints.  Our final
result will construct less straightforward bounds, again using Equation
(\ref{eq:YW_bound}) as the main assumption.  As before, assume we are implicitly
conditioning on some $\mathbf Z = \mathbf z$ everywhere.

We introduce the notation
\[
\begin{array}{rcl}
L_{xw}^{YU} & \equiv & \max(P(Y = 1 | X = x, W = w) - \epsilon_y, 0)\\
U_{xw}^{YU} & \equiv & \min(P(Y = 1 | X = x, W = w) + \epsilon_y, 1)\\
L_{w}^{XU } & \equiv & \max(P(X = 1 | W = w) - \epsilon_x, 0)\\
U_{w}^{XU}  & \equiv & \min(P(X = 1 | W = w) + \epsilon_x, 1)\\
\end{array}
\]
\noindent and define $\underline L \equiv \min\{L_{xw}^{YU}\}, \bar U \equiv \max\{U_{xw}^{YU}\}$. 
Morever, some further redundant notation is used to simplify the
description of the constraints:
\[
\begin{array}{rcl}
\delta_{1.w}^\star & \equiv & \delta_w^\star\\
\delta_{0.w}^\star & \equiv & 1 - \delta_w^\star\\
L_{11}^{XU} &\equiv& L_1^{XU}\\
L_{01}^{XU} &\equiv& 1 - U_1^{XU}\\
U_{11}^{XU} &\equiv& U_1^{XU}\\
U_{01}^{XU} &\equiv& 1 - L_1^{XU}\\
\end{array}
\]
\noindent and, following \cite{ramsahai:12}, for any $x \in \{0, 1\}$,
we define $x'$ as the complementary binary value (i.e.\ $x' =
1-x$). The same convention applies to pairs $\{w, w'\}$.
Finally, define
$\chi_{x.w} \equiv \sum_U P(X = x \ |\  W = w, U)P(U)
          = \kappa_{1x.w} + \kappa_{0x.w}$.

\begin{theorem}
\label{th:bound1}
The following constraints are entailed
by the assumptions expressed in Equations (\ref{eq:eps_y}),
(\ref{eq:eps_x}) and (\ref{eq:UW_bound2}):

\begin{equation}
\omega_{xw} \leq \min
\begin{cases}
  \kappa_{1x.w} + U_{xw}^{YU}(\kappa_{0x'.w} + \kappa_{1x'.w})\\
  \kappa_{1x.w} / L_{xw}^{XU}\\
  1 - \kappa_{0x.w} / U_{xw}^{XU}\\
\end{cases}
\label{eq:th1.1}
\end{equation}

\begin{equation}
\omega_{xw} \geq \max
\begin{cases}
  \kappa_{1x.w} + L_{xw}^{YU}(\kappa_{0x'.w} + \kappa_{1x'.w})\\
  \kappa_{1x.w} / U_{xw}^{XU}\\
  1 - \kappa_{0x.w} / L_{xw}^{XU}\\
\end{cases}
\label{eq:th1.2}
\end{equation}
\end{theorem}

\begin{theorem}
\label{th:bound2}
The following constraints are entailed
by the assumptions expressed in Equations (\ref{eq:YW_bound}), (\ref{eq:eps_y}),
(\ref{eq:eps_x}) and (\ref{eq:UW_bound2}):

\begin{equation}
\omega_{xw} \leq \min
\begin{cases}  
  (\kappa_{1x.w'} + \epsilon_w(\kappa_{0x.w'} + \kappa_{1x.w'})) / L_{xw'}^{XU}\\
  1 - (\kappa_{0x.w'} - \epsilon_w(\kappa_{0x.w'} + \kappa_{1x.w'})) / U_{xw'}^{XU}\\
\end{cases}
\label{eq:th2.1}
\end{equation}
\begin{equation}
\omega_{xw} \geq \max
\begin{cases}  
  (\kappa_{1x.w'} - \epsilon_w(\kappa_{0x.w'} + \kappa_{1x.w'})) / U_{xw'}^{XU}\\
  1 - (\kappa_{0x.w'} + \epsilon_w(\kappa_{0x.w'} + \kappa_{1x.w'})) / L_{xw'}^{XU}\\
\end{cases}
\label{eq:th2.2}
\end{equation}
\begin{equation}
\begin{array}{rcl}
\omega_{xw} - \omega_{xw'}U_{x'w}^{XU} & \leq & \kappa_{1x.w} + \epsilon_w(\kappa_{0x'.w} + \kappa_{1x'.w})\\
\omega_{xw} - \omega_{xw'}L_{x'w}^{XU} & \geq & \kappa_{1x.w} - \epsilon_w(\kappa_{0x'.w} + \kappa_{1x'.w})\\
\omega_{xw} - \omega_{xw'}U_{x'w}^{XU} & \geq & 1 - \kappa_{0x.w} - U_{x'w}^{XU} - \epsilon_w(\kappa_{0x'.w} + \kappa_{1x'.w})\\
\omega_{xw} - \omega_{xw'}L_{x'w}^{XU} & \leq & 1 - \kappa_{0x.w} - L_{x'w}^{XU} + \epsilon_w(\kappa_{0x'.w} + \kappa_{1x'.w})\\
\omega_{xw} - \omega_{xw'} & \leq & \epsilon_w\\
\omega_{xw} - \omega_{xw'} & \geq & -\epsilon_w\\
\end{array}
\label{eq:th2.3}
\end{equation}
\end{theorem}

\begin{theorem}
\label{th:bound3}
The following constraints are entailed
by the assumptions expressed in Equations (\ref{eq:YW_bound}), (\ref{eq:eps_y}),
(\ref{eq:eps_x}) and (\ref{eq:UW_bound2}):

\begin{equation}
\omega_{xw} \leq \min
\begin{cases}  
\kappa_{1x'.w'} + \kappa_{1x.w'} + \kappa_{1x.w} - \kappa_{1x'.w} + \chi_{x'w}(\bar U + \underline L + 2\epsilon_w) - \underline L\\
\kappa_{1x'.w} + \kappa_{1x.w} + \kappa_{1x.w'} - \kappa_{1x'.w'} + 2\chi_{x'w}\epsilon_w + \chi_{x'w'}(\bar U + \underline L) - \underline L\\
\end{cases}
\label{eq:th3.1}
\end{equation}
\begin{equation}
\omega_{xw} \geq \max
\begin{cases}  
- \kappa_{1x'.w'} + \kappa_{1x.w'} + \kappa_{1x'.w} + \kappa_{1x.w} + \chi_{x'w'}(\bar U + \underline L) - 2\epsilon_w\chi_{x'w} - \bar U \\
- \kappa_{1x'.w} + \kappa_{1x.w} + \kappa_{1x'.w'} + \kappa_{1x.w'} - \chi_{x'w}(2\epsilon_w - \bar U - \underline L) - \bar U \\
\end{cases}
\label{eq:th3.2}
\end{equation}
\begin{equation}
\begin{array}{rcl}
\omega_{xw} + \omega_{x'w} - \omega_{x'w'} &\geq& 
\kappa_{1x'.w} + \kappa_{1x.w} - \kappa_{1x'.w'} + \kappa_{1x.w'} -
\chi_{xw'}(\bar U + \underline L + 2\epsilon_w) + \underline L \\
\omega_{xw} + \omega_{x'w'} - \omega_{x'w} &\geq& 
\kappa_{1x'.w'} + \kappa_{1x.w'} - \kappa_{1x'.w} + \kappa_{1x.w}
- 2\chi_{xw'}\epsilon_w - \chi_{xw}(\bar U + \underline L) + \underline L\\
\omega_{xw} + \omega_{x'w'} - \omega_{x'w} &\leq& 
- \kappa_{1x'.w} + \kappa_{1x.w} + \kappa_{1x'.w'} + \kappa_{1x.w'}
- \chi_{xw}(\bar U + \underline L) + 2\epsilon_w\chi_{xw'} + \bar U \\
\omega_{xw} + \omega_{x'w} - \omega_{x'w'} &\leq& 
- \kappa_{1x'.w'} + \kappa_{1x.w'} + \kappa_{1x'.w} + \kappa_{1x.w}
+ \chi_{xw'}(2\epsilon_w - \bar U - \underline L) + \bar U \\
\end{array}
\label{eq:th3.3}
\end{equation}
\end{theorem}

Although at first sight such relations seem considerably more complex than those
given by \cite{ramsahai:12}, on closer inspection they
illustrate qualitative aspects of our free parameters. For
instance, consider
\[
\omega_{xw} \geq \kappa_{1x.w} + L_{xw}^{YU}(\kappa_{0x'.w} + \kappa_{1x'.w}),
\]
\noindent one of the instances of (\ref{eq:th1.2}). If $\epsilon_y = 1$ and $\underline \beta = \bar \beta = 1$, 
then $L_{xw}^{YU} = 0$ and this relation collapses to $\eta_{xw} \geq \zeta_{1x.w}$,
one of the original relations found by \cite{balke:97} for the standard IV model.
Decreasing $\epsilon_y$ will linearly increase $L_{xw}^{YU}$ only after
$\epsilon_y \leq P(Y = 1\ |\ X = x, W = w)$, tightening the corresponding
lower bound given by this equation. 

Consider now
\[
\omega_{xw} \leq 1 - (\kappa_{0x.w'} - \epsilon_w(\kappa_{0x.w'} + \kappa_{1x.w'})) / U_{xw'}^{XU}.
\]
\noindent If also $\epsilon_w = 0$ and $\epsilon_x = 1$, from 
this inequality it follows that $\eta_{xw} \leq 1 - \zeta_{0x.w'}$. This is another of
the standard IV inequalities \citep{balke:97}.

Equation (\ref{eq:YW_bound}) implies $|\omega_{x'w} - \omega_{x'w'}| \leq \epsilon_w$, and as such
by setting $\epsilon_w = 0$ we have that 
\begin{equation}
\label{eq:ex_4}
\omega_{xw} + \omega_{x'w} - \omega_{x'w'} \geq
\kappa_{1x'.w} + \kappa_{1x.w} - \kappa_{1x'.w'} + \kappa_{1x.w'} -
\chi_{xw'}(\bar U + \underline L + 2\epsilon_w) + \underline L 
\end{equation}
\noindent implies $\eta_{xw} \geq \eta_{1x.w} + \eta_{1x.w'} - \eta_{1x'.w'} - \eta_{0x.w'}$, one of the 
most complex relationships in \citep{balke:97}. Further geometric intuition about
the structure of the binary standard IV model is given by \cite{richardson:10a}.

These bounds are not tight, in the sense that we opt not to
fully exploit all possible algebraic combinations for some results, such
as (\ref{eq:ex_4}): there we use $\underline L \leq \eta_{xw}^\star \leq \bar
U$ and $0 \leq \delta_{w}^\star \leq 1$ instead of all possible
combinations resulting from (\ref{eq:eps_y}) and (\ref{eq:eps_x}).
The proof idea in Appendix A can be further refined,
at the expense of clarity. Because our derivation is
a further relaxation, our final bounds are more conservative
(i.e., looser).

\subsection{Efficient Optimization and Falsification Tests}
\label{sec:message_passing}

Besides providing insight into the structure of the problem, the
algebraic bounds give an efficient way of checking whether a proposed
parameter vector $\{\zeta_{yxw}\}$ is valid in Step 5 of Algorithm
\ref{tab:algo}, as well as finding the ACE bounds: we can now use
back-substitution on the symbolic set of constraints to find box
constraints $\mathcal L_{xw} \leq \omega_{xw} \leq
\mathcal U_{xw}$. The proposed parameter will be rejected
whenever an upper bound is smaller than a lower bound, and
(\ref{eq:ace}) can be trivially optimized conditioning only on the box
constraints---this is yet another relaxation, added on top of the ones
used to generate the algebraic inequalities. We 
initialize by intersecting all algebraic box constraints (of which
(\ref{eq:th1.1}) and (\ref{eq:th2.1}) are examples); next we refine these 
by scanning relations $\pm \omega_{xw} - a\omega_{xw'} \leq c$ 
(the family given by (\ref{eq:th2.3})) in lexicographical order,
and tightening the bounds of $\omega_{xw}$ using the current upper
and lower bounds on $\omega_{xw'}$ where possible. We
then identify constraints $\mathcal L_{xww'} \leq \omega_{xw} - \omega_{xw'}
\leq \mathcal U_{xww'}$ starting from $-\epsilon_w \leq \omega_{xw} -
\omega_{xw'} \leq \mathcal \epsilon_w$ and the existing bounds, and
plug them into relations $\pm\omega_{xw} + \omega_{x'w} -
\omega_{x'w'} \leq c$ (as exemplified by (\ref{eq:ex_4}))
to get refined bounds on $\omega_{xw}$ as functions of $(\mathcal L_{x'ww'}, \mathcal U_{x'ww'})$. We
iterate this until convergence, which is guaranteed since lower/upper bounds 
never decrease/increase at any iteration. This back-substitution of inequalities
follows the spirit of message-passing and it can be orders of magnitude more
efficient than the fully numerical solution, while not increasing 
the width of the intervals by too much. In Section \ref{sec:experiments}, we
provide evidence for this claim. The back-substitution method is t
used in our experiments, combined with the fully numerical linear programming approach
as explained in Section \ref{sec:experiments}. The full algorithm is given in Algorithm \ref{algo:back_sub}.

\begin{algorithm}[t]
 \SetKwInOut{Input}{input}
 \SetKwInOut{Output}{output}
 \Input{Distributions $\{\zeta_{yx.w}\}$ and $\{P(W = w)\}$;}
 \Output{Lower and upper bounds $(\mathcal L_{xw}, \mathcal U_{xw})$ for every $\omega_{xw}$}
 \BlankLine
 Find tightest lower and upper bounds $(\mathcal L_{xw}, \mathcal U_{xw})$ 
 for each $\omega_{xw}$ using inequalities (\ref{eq:th1.1}), (\ref{eq:th1.2})
 (\ref{eq:th2.1}), (\ref{eq:th2.2}), (\ref{eq:th3.1}) and (\ref{eq:th3.2})\;
 Let $\mathcal L_{xw}^{\epsilon_w}$ and $\mathcal U_{xw}^{\epsilon_w}$ be lower/upper bounds of $\omega_{xw} - \omega_{xw'}$\;
 \For{each pair $(x, w) \in \{0, 1\}^2$}{
  $\mathcal L_{xw}^{\epsilon_w} \leftarrow -\epsilon_w$\;
  $\mathcal U_{xw}^{\epsilon_w} \leftarrow \epsilon_w$\;
 }
 \While{TRUE}{
   \For {each relation $\omega_{xw} - b \times \omega_{xw'} \leq c$ in (\ref{eq:th2.3})} {
     $\mathcal U_{xw}^{\epsilon_w} \leftarrow \min\{\mathcal U_{xw}^{\epsilon_w}, (b - 1)\mathcal L_{xw} + c\}$
   }
   \For {each relation $\omega_{xw} - b \times \omega_{xw'} \geq c$ in (\ref{eq:th2.3})} {
     $\mathcal L_{xw}^{\epsilon_w} \leftarrow \max\{\mathcal L_{xw}^{\epsilon_w}, (b - 1)\mathcal U_{xw} + c\}$
   }
   \For {each relation $\omega_{xw} + \omega_{x'w} - \omega_{x'w'} \leq c$ in (\ref{eq:th3.3})} {
     $\mathcal U_{xw} \leftarrow \min\{\mathcal U_{xw}, c - \mathcal L_{xw'}^{\epsilon_w}\}$
   }
   \For {each relation $\omega_{xw} - (\omega_{x'w} - \omega_{x'w'}) \leq c$ in (\ref{eq:th3.3})} {
     $\mathcal U_{xw} \leftarrow \min\{\mathcal U_{xw}, c + \mathcal U_{xw'}^{\epsilon_w}\}$
   }
   \For {each relation $\omega_{xw} + \omega_{x'w} - \omega_{x'w'} \geq c$ in (\ref{eq:th3.3})} {
     $\mathcal U_{xw} \leftarrow \max\{\mathcal U_{xw}, c - \mathcal U_{xw'}^{\epsilon_w}\}$
   }
   \For {each relation $\omega_{xw} - (\omega_{x'w} - \omega_{x'w'}) \geq c$ in (\ref{eq:th3.3})} {
     $\mathcal U_{xw} \leftarrow \max\{\mathcal U_{xw}, c + \mathcal L_{xw'}^{\epsilon_w}\}$
   }
   \If{no changes in $\{(\mathcal L_{xw}, \mathcal U_{xw})\}$}{
     break
   }
 }
 \Return{$(\mathcal L_{xw}, \mathcal U_{xw})$ for each $(x, w) \in \{0, 1\}^2$}
 \BlankLine
 \caption{The iterative back-substitution procedure for
   bounding $\mathcal L_{xw} \leq \omega_{xw} \leq \mathcal U_{xw}$
   for all combinations of $x$ and $w$ in $\{0, 1\}^2$.}
\label{algo:back_sub}
\end{algorithm}

\section{Choosing Relaxation Parameters}
\label{sec:parameters}

The free parameters $\aleph \equiv \{\epsilon_w, \epsilon_x, \epsilon_y,
\underline \beta, \bar \beta\}$ do not have an unique, clear-cut,
domain-free procedure by which they can be calibrated. However,
as we briefly discussed in Section \ref{sec:wpp}, it is 
useful to state explicitly the following worst-case scenario
guarantee of {\sc WPP}:

\begin{corollary}
\label{coll:faithfulness}
Given $W \ncind Y\ |\ \mathbf Z$ and
$W \cind Y\ |\ \{X, \mathbf Z\}$, the {\sc WPP} population bounds on the ACE will always
include the back-door adjusted population ACE based on $\mathbf Z$.
\end{corollary}

\noindent {\bf Proof} The proof follows directly by plugging in
the quantities $\epsilon_w = \epsilon_y = \epsilon_x = 0$, $\underline
\beta = \bar \beta = 1$, into the analytical bounds of Section \ref{sec:bounds}, which will give the
tightest bounds on the ACE (generalized to accommodate a background set
$\mathbf Z$): a single point, which also happens to be the functional
obtained by the back-door adjustment. \hfill\BlackBox\\

The implication is that, regardless of the choice of free parameters,
the result is guaranteed to be more conservative than the one obtained
using the faithfulness assumption. In any case, this does not mean that
a judicious choice of relaxation parameters is of secondary importance.

The setting of relaxation parameters can be interpreted in two ways:
\begin{itemize}
\item $\aleph$ is set prior to calculating the ACE; this uses expert knowledge concerning the remaining amount of unmeasured confounding, decided
with respect to the provided admissible set and witness, or by a default rule concerning beliefs on faithfulness violations;
\item $\aleph$ is deduced by the outcome of a sensivity analysis procedure; given a particular interval length $L$, we derive a quantification
of faithfulness violations (represented by $\aleph$) required to
generate causal models compatible with the observational data and an interval of length $L$
containing the ACE;
\end{itemize}

That is, in the first scenario the input is $\aleph$, the output are
bounds on the ACE. In the second scenario, the input is the acceptable
width of an interval containing the ACE, the output are the bounds on
the ACE {\it and} a choice of $\aleph$. In his rejoinder to the discussion of
\citep{rosenbaum:02}, Rosenbaum points out that the sensitivity
analysis procedure just states the logical outcome of the structural
assumptions: the resulting deviation of, say, $P(Y = 1\ |\ X = x, W =
w)$ from $P(Y = 1\ |\ X = x, W = w, U = u)$ required to explain the
given length of variation on the ACE is not directly imposed by expert
knowledge concerning confounding effects. Expert knowledge is of
course still necessary to decide whether the resulting deviation is
unlikely or not (and hence, whether the resulting interval is
believable), although communication by sensitivity analysis might
facilitate discussion and criticism of the study.

Motivated by the idea of starting from a pre-specified length $L$ for
the resulting interval around the ACE, in what follows we describe two
possible ways of setting relaxation parameters. We contrast the methods against the idea of putting priors on
latent variable models, as discussed in Section
\ref{sec:unidentifiability}.

\subsection{Choice by Grid Search Conditioned on Acceptable Information Loss}
\label{sec:tradeoff}

One pragmatic default rule is to first ask how wide an ACE interval
can be so that the result is still useful for the goals of the
analysis (e.g., sorting possible controls $X$ as candidates for a lab
experiment based on lower bounds on the ACE). Let $L$ be the interval
width the analyst is willing to pay for. Set $\epsilon_w = \epsilon_x
= \epsilon_y = k_\epsilon$ and $\underline \beta = c$, $\bar \beta = 1 / c$,
for some pair $(k_\epsilon, c)$ such that $0 \leq k < 1$, $0 < c \leq 1$, and
let $(k, c)$ range over a grid of values. For each witness/admissible
set candidate pair, pick the $(k, c)$ choice(s) entailing interval(s)
of length closest to $L$. In case of more than one solution, summarize
them by a criterion such the union of the intervals.

This methodology provides an explicit trade-off between length of the
interval and tightness of assumptions. Notice that, starting from the
backdoor-adjusted point estimator of \cite{entner:13}, it is not clear
how one would build a procedure to provide such a trade-off: that is,
a procedure by which one could build an interval around the point estimate
within a given acceptable amount of information loss. {\sc WPP} provides a
principled way of building such an interval, with the resulting
assumptions on $\aleph$ being explicitly revealed as a by-product.  If
the analyst believes that the resulting values of $\aleph$ are not
strict enough, and no substantive knowledge exists that allows
particular parameters to be tightened up, then one either has to
concede that wider intervals are necessary or to find other means of
identifying the ACE unrelated to the faithfulness assumption. 

In the experiments in Section \ref{sec:synth}, we define a parameter space of $k_\epsilon
\in \{0.05, 0.10, \dots, 0.30\}$ and $c \in \{0.9, 1\}$. More than one
interval of approximately the same width are identified. For instance,
the configurations $(k_\epsilon = 0.25, c = 1)$ and $(k_\epsilon =
0.05, c = 0.9)$ both produce intervals of approximately length $0.30$.

\subsection{Linking Selection on the Observables to Selection on the Unobservables}
\label{sec:selection}

The trade-off framework assumes the analyst has a known tolerance
level for information loss (that is, the length of the interval around
the back-door adjusted estimator), around which an automated procedure
for choosing $\aleph$ can be constructed. Alternatively, one might
choose a value of $\aleph$ \emph{a priori} using information from the
problem at hand, and accept the information loss that it entails. 
This still requires a way of connecting prior assumptions to data.

Observational studies cannot be carried out without making assumptions
that are untestable given the data at hand. There will always be
degrees of freedom that must be chosen, even if such choices are open to
criticism. The game is to provide a language to express
assumptions in as transparent a manner as possible. Our view on
priors for the latent variable model (Section
\ref{sec:unidentifiability}) is that such prior knowledge is far too
difficult to justify when the interpretation of $U$ is
unclear. Moreover, putting a prior on a parameter such as $P(Y = 1\ |
X = x, W = w, U = u)$ so that this prior is bounded by the constraint
$|P(Y = 1\ | X = x, W = w, U = u) - P(Y = 1\ |\ X = w, W = w)| \leq
\epsilon_w$ has no clear advantage over the {\sc WPP}: a specification of
the shape of this prior is still necessary and may have undesirable
side effects; it has no computational advantages over the {\sc WPP}, as
constraints will have to be dealt with now within a Markov chain Monte
Carlo procedure; it provides no insight on how constraints are related to one another 
(Section \ref{sec:scale-up}); it still suggests a point estimate that
should not be trusted lightly, and posterior bounds which cannot be
interpreted as worst-case bounds; and it still requires a choice of
$\epsilon_w$.

That is not to say that subjective priors on the relationship between
$U$ and the observables cannot be exploited, but the level of
abstraction at which they need to be specified should have
advantages when compared to the latent variable model approach. For
instance, \cite{altonji:05} introduced a framework to deal with
violations of the IV assumptions (in the context of linear
models). Their main idea is to linearly decompose the (observational)
dependence of $W$ and $\mathbf Z$, and the (causal) dependence of $Y$
and $\mathbf Z$, as two signal-plus-noise decompositions, and assume
that dependence among the signals allows one to infer the dependence
among the noise terms. In this linear case, the dependence among noise
terms gives the association between $W$ and $Y$ through unmeasured
confounders. The constraint given by the assumption can then be used
to infer bounds on the (differential) ACE. The details are not
straightforward, but the justification for the assumption is
indirectly derived by assuming $\mathbf Z$ is chosen by a sampling
mechanism that picks covariates from the space of confounders $U$, so
that $|\mathbf Z|$ and $|U|$ are large. The principal idea is that the
dependence between the covariates which are observed (i.e.\ $\mathbf Z$) 
and the other variables ($W, X, Y$) 
should tell us something about the impact of the unmeasured
confounders. Their method is presented for linear models only, and the
justification requires a very large $|\mathbf Z|$.

We introduce a very different method inspired by the same general
principle, but exploiting the special structure of our procedure.
Instead of relying on linearity and a fixed set of covariates,
consider the following postulate: the variability of back-door
adjusted ACE estimators based on different admissible sets, as implied
by Rule 1, should provide some information about the extent of the
violations of faithfulness in the given domain.

For simplicity of exposition, we adopt the parameterization of
$\aleph$ as given by the three parameters $(\epsilon_w, \epsilon_{xy}
= \epsilon_x = \epsilon_y, \beta = \underline \beta = 1 / \bar
\beta)$. Given a prior $\pi(\epsilon_w, \epsilon_{xy}, \beta)$ over the three-dimensional
unit cube $[0, 1]^3$, we want to assess probable values of such
parameters using a ``likelihood'' function that explains the
variability of the ACEs provided by Entner et al.'s rule. We want the
posterior to converge to the single values $\epsilon_w = 0,
\epsilon_{xy} = 0$ and $\beta = 1$ as the number of witness/admissible
set pairs increase and under the condition that they agree on the same
value.

For that, we will consider a {\it target} witness/admissible set pair
$(W^\star, \mathbf Z^\star)$, and a {\it reference set} $\mathcal R$
of other admissible sets. Given $\aleph$, $W^\star,
\mathbf Z^\star$ and the joint distribution over observables
$P(\mathbf V)$, a lower bound $LB^\star$ and an upper bound $UB^\star$ on the ACE 
are determined. Given the bounds, we define a likelihood function
\begin{equation}
\label{eq:lik_aleph}
\mathcal L(\aleph; P(\mathbf V), W^\star, \mathbf Z^\star, \mathcal R) \equiv
\prod_{i = 1}^n p_{N[-1, 1]}(ACE_i; m(LB^\star, UB^\star), v(LB^\star, UB^\star))
\end{equation}
\noindent where $p_{N[-1, 1]}(\cdot; m, v)$ is a truncated Gaussian density on $[-1, 1]$ proportional
to a Gaussian with mean $m$ and variance $v$; $m(LB^\star, UB^\star)$
and $v(LB^\star, UB^\star)$ are functions of the bounds. Along with
the prior, this defines a posterior over $\aleph$; $ACE_i$ is the
back-door adjusted ACE obtained with the $i$th entry of $\mathcal R$, conditioned on $P(\mathbf V)$.

There are many degrees of freedom in this formulation, and we do not
claim it represents anything other than subjective knowledge: an
approximation to the idea that large/small variability of the ACEs
should indicate large/small violations of faithfulness, and that we
should get more confident about the magnitude of the violations as
more ACEs are reported by Entner et al.'s back-door estimator. In our
implementation we treat $m$ as a free parameter, with a uniform
prior in $[LB^\star, UB^\star]$. We treat $v$ as a deterministic
function of the bounds,
\begin{equation}
v(LB^\star, UB^\star) \equiv ((UB^\star - LB^\star) / 6)^2
\end{equation}
\noindent to reflect the assumption that the interval $[LB^\star, UB^\star]$ 
should cover a large amount of mass of the model---in this case,
$UB^\star - LB^\star$ is approximately 6 times the standard deviation
of the likelihood model. 

Finally, our problem has one last degree of freedom:
(\ref{eq:lik_aleph}) treats the ACEs implied by $\mathcal R$ as
conditionally independent. Since many admissible sets overlap, this
can result in overconfident posteriors, in the sense that they do not
reflect our belief that similar admissible sets do not provide
independent pieces of evidence concerning violations of
faithfulness. Our pragmatic correction to that is to discard from
$\mathcal R$ any admissible set which is a strict superset of some
other element of $\mathcal R \cup \{\mathbf Z^\star\}$. Notice that in some
situations, $\mathcal R$ might contain the empty set as a possible
admissible set, implying that the resulting $\mathcal R$ will contain
at most one element (the empty set itself). Optionally, one might forbid 
\emph{a priori}
the empty set ever entering $\mathcal R$. 

The criterion above can be refined in many ways: among other issues,
one does not want to inflate the confidence on $\aleph$ by measuring
many highly correlated (sets of) covariates that will end up being
added independently to $\mathcal R$. One idea is to modify the
likelihood function to allow for dependencies among different ACE
``data points.'' We leave this as future work.

Besides the priors over $\aleph$ and $m(LB^\star, UB^\star)$, we can also in
principle define a prior for $P(\mathbf V)$. In the following illustration,
and in the application in Section \ref{sec:influenza}, we simplify the analysis 
by treating $P(\mathbf V)$ as known, using the posterior expected value of
$P(\mathbf V)$ given an BDeu prior with effective sample size of 10.
The full algorithm is shown in Algorithm \ref{algo:sel_aleph}.

\begin{algorithm}[t]
 \SetKwInOut{Input}{input}
 \SetKwInOut{Output}{output}
 \Input{Data set $\mathcal D$ over observed variables $\mathbf V$; hyperparameter $\alpha$ for the BDeu prior;
  prior $\pi(\aleph)$; a flag {\tt allow\_empty} indicating whether empty sets are allowed}
 \Output{A posterior distribution over $\aleph$}
 \BlankLine

 Find all witness/admissible set pairs $\mathcal P$ according to Rule 1, data $\mathcal D$ and BDeu hyperparameter $\alpha$
 
 Let $(W^\star, \mathbf Z^\star)$ be the highest scoring pair according to the 
 {\sc WPP} scoring rule (\ref{eq:wpp_score})

 Let $\mathcal R$ be the set of all admissible sets in $\mathcal P$

 Remove the empty set from $\mathcal R$ if {\tt allow\_empty} is false
 
 Remove from $\mathcal R$ any set that strictly contains some other set in $\mathcal R$

 Remove $\mathbf Z^\star$ from $\mathcal R$

 Let $P(\mathbf V)$ be the posterior expected value of the distribution of $\mathcal V$ as
 given by $\mathcal D$ and $\alpha$

 Return the posterior distribution implied by $\pi(\aleph)$, $\mathcal R$,
 $(W^\star, \mathbf Z^\star)$ and $P(\mathbf V)$

 \BlankLine
\caption{Finding a posterior distribution over relaxation parameters $\aleph$ using candidate solutions
 generated by Entner et al.'s Rule 1.}
\label{algo:sel_aleph}
\end{algorithm}

It should be stressed out that the posterior over $\aleph$ will in
general be unidentifiable, since the sufficient statistics for the
likelihood are the upper and lower bounds and
different values of $\aleph$ can yield the same bounds. Our
implementation of Algorithm \ref{algo:sel_aleph} consists of using a
simple Metropolis-Hastings scheme to sample each of the three
components $\epsilon_w, \epsilon_{xw}, \beta$ one at a time, and mixing
will be a practical issue. Priors will matter. In particular, a situation with a very small
$\mathcal R$ and an uniform prior $\pi(\aleph)$ might require many MCMC iterations. 

\begin{figure}[t]
\begin{center}
\begin{tabular}{cc}
\includegraphics[width=3in,height=3in]{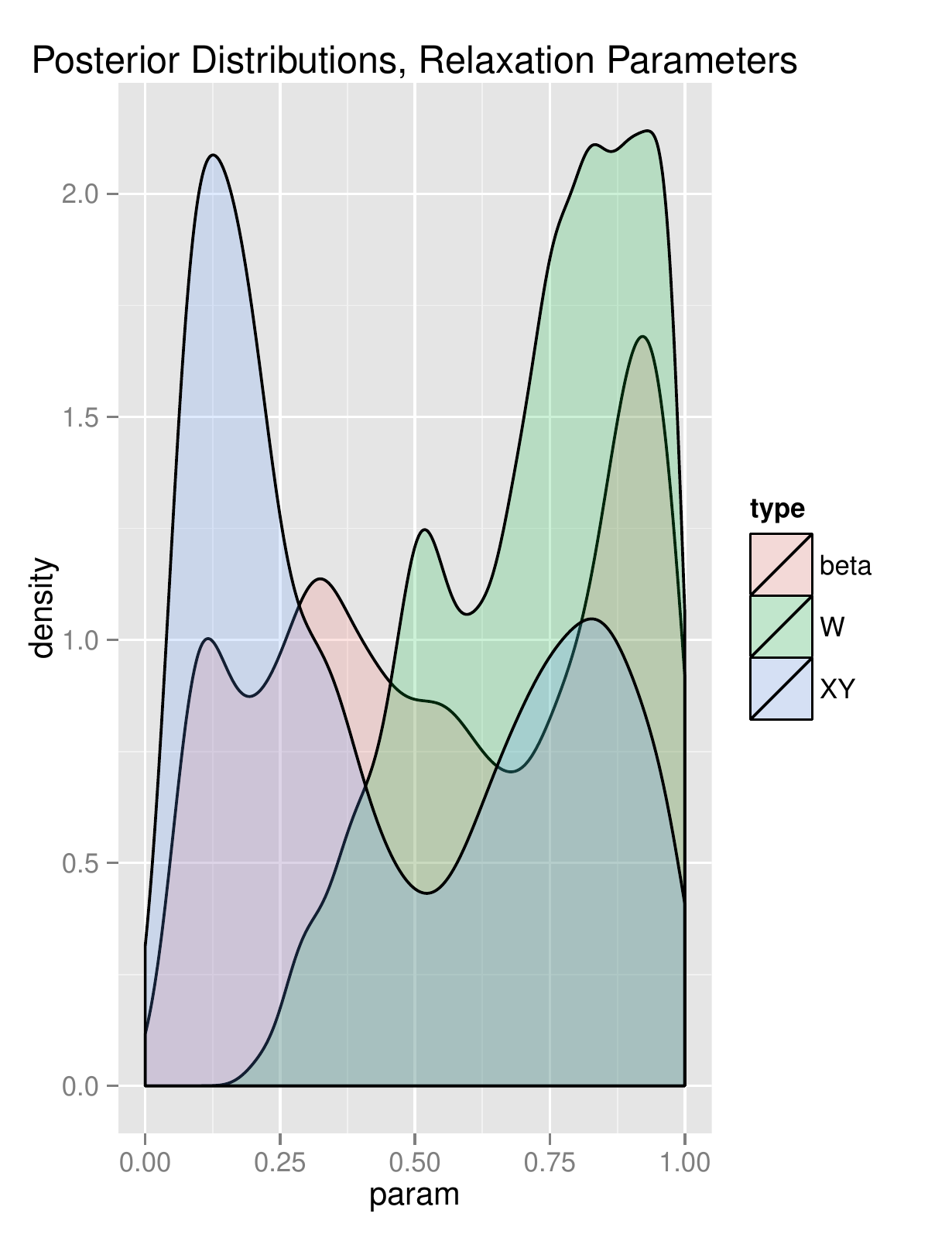} &
\includegraphics[width=3in,height=3in]{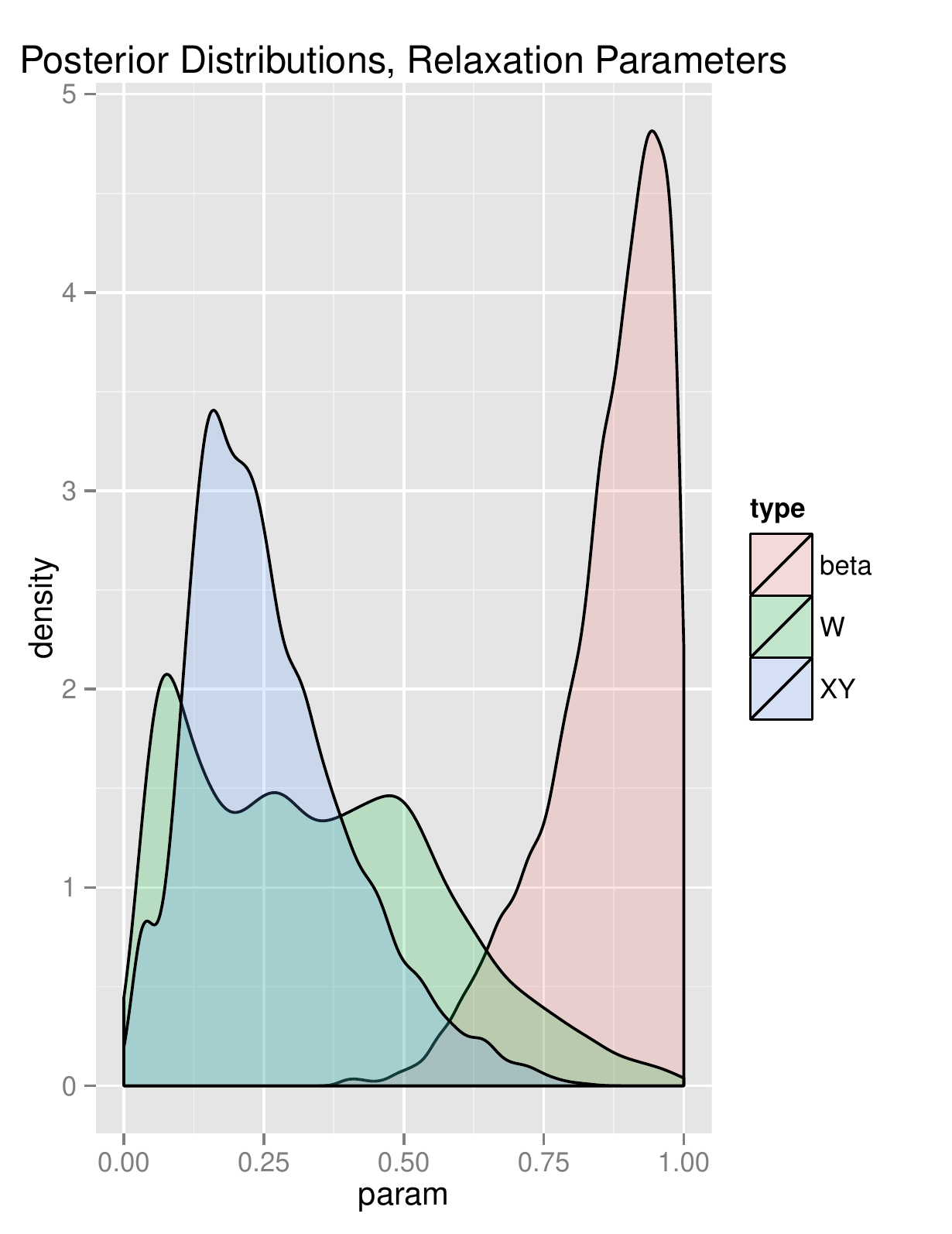} \\
(a) & (b)\\
\end{tabular}
\end{center}
\caption{In (a), the posterior marginal densities for $\epsilon_w, \epsilon_{xy}$ and $\beta$ with an uniform prior,
smoothed kernel density estimates based on $10,000$ Monte Carlo
samples. An analogous picture is shown in (b), for the situation where
the prior is now a product of three univariate truncated Gaussians in $[0, 1]$, each
marginal proportional to an univariate Gaussian with means $(0.2, 0.2,
0.95)$ and variances $(0.1, 0.1, 0.05)$, respectively.}
\label{fig:aleph_inference_example}
\end{figure}

Consider Figure \ref{fig:aleph_inference_example}. Here, we have a
synthetic problem where we know no admissible set exists. Due to
sampling variability and near-faithfulness violations, {\sc WPP}
identifies three such sets.  This is one of the hardest positions
for the $\aleph$ learning procedure, since the posterior will also be
very broad.  The true ACE is $-0.16$, while the estimated ACEs given
by $\mathcal R$ are $\{-0.44, -0.34\}$. With only two (reasonably
spread out) data points and an uniform prior for $\aleph$, we obtain
the posterior distribution for the entries of $\aleph$ as shown in
Figure
\ref{fig:aleph_inference_example}(a). This reflects uncertainty and
convergence difficulties of the MCMC procedure.  More informative
priors make a difference, as shown in Figure
\ref{fig:aleph_inference_example}(b). In Section \ref{sec:influenza},
a simple empirical study with far more concentrated ACEs provides a far
more tightly concentrated set of marginal posteriors.

\section{Experiments}
\label{sec:experiments}

In this Section, we start with a comparison of the back-substitution
algorithm of Section \ref{sec:message_passing} against the fully
numerical procedure, which generates constraints using standard
algorithms for changing between polytope representations. We then
perform studies with synthetic data, comparing different back-door
estimation algorithms against {\sc WPP}. Finally, we perform analysis
with a real dataset.

\subsection{Empirical Investigation of the Back-substitution Algorithm}

We compare the back-substitution algorithm introduced in Section
\ref{sec:message_passing} with the fully numerical
algorithm. Comparison is done in two ways:
(i) computational cost, as measured by the wallclock time taken to
generate 100 samples by rejection sampling; (ii) width of the
generated intervals. As discussed in Section
\ref{sec:message_passing}, bounds obtained by the back-substitution
algorithm are at least as wide as in the numerical algorithm, barring
rounding problems\footnote{About $1\%$ of the time we observed numerical problems
with the polytope generator, as we were not using rational arithmetic in order to
speed it up. Those were excluded from the statistics reported in this Section.}. 

We ran two batches of 1000 trials each, varying the level of the
relaxation parameters. In the first batch, we set $\epsilon_x =
\epsilon_y = \epsilon_w = 0.2$, and $\underline \beta = 0.9$, $\bar
\beta = 1.1$. In the second batch, we change parameters so that
$\underline \beta = \bar \beta = 1$. Experiments were run on a
Intel Xeon E5-1650 at 3.20Ghz. Models were simulated according the the
structure $W \rightarrow X \rightarrow Y$, sampling each conditional
distribution of a vertex being equal to 1 given its parent from the
uniform $(0, 1)$ distribution. The numerical procedure of converting
extreme points to linear inequalities was done using the package {\sc
rcdd}, a {\sc R} wrapper for the {\it cddlib} by Komei
Fukuda. Inference is done by rejection sampling, requiring 100 samples
per trial. We fix the number of interations of the back-substitution
method to 4, which is more than enough to achieve convergence.
All code was written in {\sc R}.

For the first batch, the average time difference between the fully
numerical method and the back-substitution algorithm was 1 second, standard
deviation (s.d.) 0.34. The ratio between times had a mean of 203
(s.d. 82). Even with a more specialized implementation of the
polytope dualization step\footnote{One advantage of the analytical
bounds, as used by the back substitution method, is that it is easy to
express them as matrix operations over all Monte Carlo samples, while
the polytope construction requires iterations over the samples.}, two
orders of magnitude of difference seem hard to remove by better
coding. Concerning interval widths, the mean difference was 0.15
(s.d. 0.06), meaning that the back-substitution on average has
intervals where the upper bound minus the lower bound difference is
0.15 units more than the numerical method, under this choice of
relaxation parameters and averaged over problems generated according
to our simulation scheme. There is a correlation between the width
difference and the interval width given by the numerical method the
gap, implying that differences tend to be larger when bounds are
looser: the gap between methods was as small as 0.04 for a fully
numerical interval of width 0.19, and as large as 0.23 for a fully
numerical interval of width 0.49. For the case where $\bar \beta =
\underline \beta = 1$, the average time difference was 0.92 (s.d. of
0.24), ratio of 152 (s.d. 54.3), interval width difference of 0.09
(s.d. 0.03); The gap was as small as 0.005 for a fully numerical
interval of width 0.09, and as large as 0.17 for a fully numerical
interval of with 0.23.

\subsection{Synthetic Studies}
\label{sec:synth}

We describe a set of synthetic studies where we assess the trade-off
between ACE intervals and error, as wider intervals will be less
informative than point estimators such as the back-door adjustment,
but by definition have more chances of correctly covering the ACE.

In the synthetic study setup, we compare our method against NE1 and
NE2, two na\"ive point estimators defined by back-door adjustment on
the whole of set of available covariates $\mathcal W$ and on the empty
set, respectively. The former is widely used in practice, even when
there is no causal basis for doing so \citep{pearl:09a}. The point
estimator of \cite{entner:13}, based solely on the faithfulness
assumption, is also assessed.

We generate problems where conditioning on the whole set $\mathcal W$
is guaranteed to give incorrect estimates. In detail: we
  generate graphs where $\mathcal W \equiv \{Z_1, Z_2, \dots,
  Z_8\}$. Four independent latent variables $L_1, \dots, L_4$ are
  added as parents of each $\{Z_5, \dots, Z_8\}$; $L_1$ is also a
  parent of $X$, and $L_2$ a parent of $Y$. $L_3$ and $L_4$ are each
  randomly assigned to be a parent of either $X$ or $Y$, but not both.
  $\{Z_5, \dots, Z_8\}$ have no other parents. The graph over $Z_1,
  \dots, Z_4$ is chosen by adding edges uniformly at random according
  to the lexicographic order. In consequence using the full set
  $\mathcal W$ for back-door adjustment is always incorrect, as at
  least four paths $X \leftarrow L_1 \rightarrow Z_i \leftarrow L_2
  \rightarrow Y$ are active for $i = 5, 6, 7, 8$. The conditional
  probabilities of a vertex given its parents are generated by a
  logistic regression model with pairwise interactions, where
  parameters are sampled according to a zero mean Gaussian with
  standard deviation 20 / number of parents. Parameter values are also
  squashed, so that if the generated value if greater than $0.975$ or 
  less than $0.025$, it is resampled uniformly in $[0.950, 0.975]$ or
  $[0.025, 0.050]$, respectively.

We analyze two variations: one where it is guaranteed that at least
one valid pair witness-admissible set exists; in the other, all latent
variables in the graph are set also as common parents also of $X$ and
$Y$, so no valid witness exists. We divide each variation into two
subcases: in the first, ``hard'' subcase, parameters are chosen (by
rejection sampling, proposing from the model described in the previous
paragraph) so that NE1 has a bias of at least 0.1 in the population;
in the second, no such a selection exists, and as such our exchangeable
parameter sampling scheme makes the problem relatively easy. We
summarize each {\sc WPP} interval by the posterior expected value of
the lower and upper bounds. In general {\sc WPP} returns more than one
bound: we select the upper/lower bound corresponding to the $(W,
\mathbf Z)$ pair which maximizes the score described at the end of
Section \ref{sec:bayes_learn}.  A BDeu prior with an equivalent sample size of $10$
was used.

Our main evaluation metric for an estimate is the Euclidean distance
(henceforth, ``error'') between the true ACE and the closed point in
the given estimate, whether the estimate is a point or an
interval. For methods that provide point estimates (NE1, NE2, and
faithfulness), this means just the absolute value of the difference
between the true ACE and the estimated ACE. For {\sc WPP}, the error
of the interval $[\mathcal L, \mathcal U]$ is zero if the true ACE
lies in this interval. We report {\it error average} and {\it error
tail mass at 0.1}, the latter meaning the proportion of cases where
the error exceeds 0.1. Moreover, the faithfulness estimator is defined
by averaging over all estimated ACEs as given by the
accepted admissible sets in each problem.

As discussed in Section \ref{sec:tradeoff}, {\sc WPP} can be
understood as providing a trade-off between information loss and
accuracy.  For instance, while the trivial interval $[-1, 1]$ will
always have zero error, it is not an interesting solution. We assess
the trade-off by running simulations at different levels of
$k_\epsilon$, where $\epsilon_w = \epsilon_y = \epsilon_x =
k_\epsilon$.  We also have two configurations for $\{\underline \beta,
\bar \beta\}$: we set them at either $\underline \beta = \bar \beta =
1$ or $\underline \beta = 0.9, \bar \beta = 1.1$.

For the cases where no witness exists, Entner's Rule 1 should
theoretically report no solution. \cite{entner:13} used stringent
thresholds for deciding when the two conditions of Rule 1 held.
Instead we take a more relaxed approach, using a uniform
prior on the hypothesis of independence.  As such, due to the nature
of our parameter randomization, more often than not is will propose at
least one witness. That is, for the problems where no exact solution
exists, we assess how sensitive the methods are given conclusions
taken from ``approximate independencies'' instead of exact ones.

The analytical bound are combined with the numerical procedure as
follows.  We use the analytical bounds to test each proposed model
using the rejection sampling criterion. Under this scheme, we
calculate the posterior expected value of the contingency table and,
using this single point, calculate the bounds using the fully
numerical method. This is not guaranteed to work: the point estimator
using the analytical bounds might lie outside the polytope given by
the full set of constraints. If this situation is detected, we revert
to calculating the bounds using the analytical method. The gains in
interval length reduction using the full numerical method are relatively
modest (e.g., at $k_\epsilon = 0.20$, the average interval width
reduced from $0.30$ to $0.24$) but depending on the application they
might make a sensible difference.


\begin{table}
\begin{center}
\begin{tabular}{c|c||c|c||c|c||c||c|c||c}
\hline
\multicolumn{10}{l}{{\bf Hard, Solvable:} NE1 = $(0.12, 1.00)$, NE2 = $(0.02, 0.03)$}\\
\hline
$k_\epsilon$ & Found & \multicolumn{2}{c||}{Faith.1} & \multicolumn{2}{c||}{{\sc WPP}1} & Width1 
                                                     & \multicolumn{2}{c||}{{\sc WPP}2} & Width2 \\
\hline
$0.05$ & $0.74$ & $0.03$ & $0.05$ & $0.02$ & $0.05$ & $0.05$ & $0.00$ & $0.00$ & $0.34$\\ 
$0.10$ & $0.94$ & $0.04$ & $0.05$ & $0.01$ & $0.01$ & $0.11$ & $0.00$ & $0.00$ & $0.41$\\ 
$0.15$ & $0.99$ & $0.04$ & $0.05$ & $0.01$ & $0.02$ & $0.16$ & $0.00$ & $0.00$ & $0.46$\\ 
$0.20$ & $1.00$ & $0.05$ & $0.05$ & $0.01$ & $0.01$ & $0.24$ & $0.00$ & $0.00$ & $0.53$\\ 
$0.25$ & $1.00$ & $0.05$ & $0.07$ & $0.00$ & $0.00$ & $0.32$ & $0.00$ & $0.00$ & $0.60$\\ 
$0.30$ & $1.00$ & $0.05$ & $0.10$ & $0.00$ & $0.00$ & $0.41$ & $0.00$ & $0.00$ & $0.69$\\ 
\hline
\multicolumn{10}{l}{{\bf Easy, Solvable:} NE1 = $(0.01, 0.01)$, NE2 = $(0.07, 0.24)$}\\
\hline
$k_\epsilon$ & Found & \multicolumn{2}{c||}{Faith.1} & \multicolumn{2}{c||}{{\sc WPP}1} & Width1 
                                                     & \multicolumn{2}{c||}{{\sc WPP}2} & Width2 \\
\hline
$0.05$ & $0.81$ & $0.03$ & $0.02$ & $0.02$ & $0.04$ & $0.04$ & $0.00$ & $0.01$ & $0.34$\\ 
$0.10$ & $0.99$ & $0.02$ & $0.02$ & $0.01$ & $0.02$ & $0.09$ & $0.00$ & $0.00$ & $0.40$\\ 
$0.15$ & $1.00$ & $0.02$ & $0.01$ & $0.00$ & $0.00$ & $0.17$ & $0.00$ & $0.00$ & $0.46$\\ 
$0.20$ & $1.00$ & $0.02$ & $0.01$ & $0.00$ & $0.00$ & $0.24$ & $0.00$ & $0.00$ & $0.54$\\ 
$0.25$ & $1.00$ & $0.02$ & $0.01$ & $0.00$ & $0.00$ & $0.32$ & $0.00$ & $0.00$ & $0.61$\\ 
$0.30$ & $1.00$ & $0.02$ & $0.01$ & $0.00$ & $0.00$ & $0.41$ & $0.00$ & $0.00$ & $0.67$\\
\hline
\multicolumn{10}{l}{{\bf Hard, Not Solvable:} NE1 = $(0.16, 1.00)$, NE2 = $(0.20, 0.88)$}\\
\hline
$k_\epsilon$ & Found & \multicolumn{2}{c||}{Faith.1} & \multicolumn{2}{c||}{{\sc WPP}1} & Width1 
                                                     & \multicolumn{2}{c||}{{\sc WPP}2} & Width2 \\
\hline
$0.05$ & $0.67$ & $0.20$ & $0.90$ & $0.17$ & $0.76$ & $0.06$ & $0.04$ & $0.14$ & $0.32$\\ 
$0.10$ & $0.91$ & $0.19$ & $0.91$ & $0.13$ & $0.63$ & $0.10$ & $0.02$ & $0.07$ & $0.39$\\ 
$0.15$ & $0.97$ & $0.19$ & $0.92$ & $0.10$ & $0.41$ & $0.18$ & $0.01$ & $0.03$ & $0.45$\\ 
$0.20$ & $0.99$ & $0.19$ & $0.95$ & $0.07$ & $0.25$ & $0.24$ & $0.01$ & $0.01$ & $0.51$\\ 
$0.25$ & $1.00$ & $0.19$ & $0.96$ & $0.03$ & $0.13$ & $0.31$ & $0.00$ & $0.00$ & $0.58$\\ 
$0.30$ & $1.00$ & $0.19$ & $0.96$ & $0.02$ & $0.06$ & $0.39$ & $0.00$ & $0.00$ & $0.66$\\
\hline
\multicolumn{10}{l}{{\bf Easy, Not Solvable:} NE1 = $(0.09, 0.32)$, NE2 = $(0.14, 0.56)$}\\
\hline
$k_\epsilon$ & Found & \multicolumn{2}{c||}{Faith.1} & \multicolumn{2}{c||}{{\sc WPP}1} & Width1 
                                                     & \multicolumn{2}{c||}{{\sc WPP}2} & Width2 \\
\hline
$0.05$ & $0.68$ & $0.13$ & $0.51$ & $0.10$ & $0.37$ & $0.05$ & $0.02$ & $0.07$ & $0.33$\\ 
$0.10$ & $0.97$ & $0.12$ & $0.53$ & $0.08$ & $0.28$ & $0.10$ & $0.01$ & $0.05$ & $0.39$\\ 
$0.15$ & $1.00$ & $0.12$ & $0.52$ & $0.05$ & $0.17$ & $0.16$ & $0.01$ & $0.03$ & $0.46$\\ 
$0.20$ & $1.00$ & $0.12$ & $0.53$ & $0.03$ & $0.08$ & $0.23$ & $0.01$ & $0.03$ & $0.52$\\ 
$0.25$ & $1.00$ & $0.12$ & $0.48$ & $0.02$ & $0.05$ & $0.31$ & $0.00$ & $0.02$ & $0.59$\\ 
$0.30$ & $1.00$ & $0.12$ & $0.48$ & $0.01$ & $0.04$ & $0.39$ & $0.00$ & $0.01$ & $0.65$\\ 
\end{tabular}
\end{center}
\label{tab:synth}
\caption{Summary of the outcome of the synthetic studies. 
   Columns labeled {\sc WPP}1 refer to results obtained
  for $\underline \beta = \bar \beta = 1$, while {\sc WPP}2 refers to
  the case $\underline \beta = 0.9, \bar \beta = 1.1$. The first
  column is the level in which we set the remaining parameters,
  $\epsilon_x = \epsilon_y = \epsilon_w = k_\epsilon$. The second
  column is the frequency by which a {\sc WPP} solution has been found
  among $100$ runs. For each particular method (NE1, NE2, Faithfulness
  and {\sc WPP}) we report the pair (error average, error tail mass at
  0.1), as explained in the main text. The Faithfulness estimator is
  the back-door adjustment obtained by using as the admissible set the
  same set found by {\sc WPP}1. Averages are taken only over the cases
  where a witness-admissible set pair has been found. The columns
  following each {\sc WPP} results are the median width of the
  respective {\sc WPP} interval across the $100$ runs.}
\end{table}

We simulate 100 datasets for each one of the four cases (hard
case/easy case, with theoretical solution/without theoretical
solution), 5000 points per dataset, 1000 Monte Carlo samples per
decision. Results for the point estimators (NE1, NE2, faithfulness)
are obtained using the population contingency tables. Results are
summarized in Table \ref{tab:synth}. The first observation is at very
low levels of $k_\epsilon$ we increase the ability to reject all
witness candidates: this is due mostly not because Rule 1 never fires,
but because the falsification rule of {\sc WPP} (which does not
enforce independence constraints) rejects the proposed witnesses found
by Rule 1. The trade-off set by {\sc WPP} is quite stable, where
larger intervals are indeed associated with smaller error. The point
estimates vary in quality, being particularly bad in the situation
where no witness should theoretically exist.  The set-up where
$\underline \beta = 0.9, \bar \beta = 1$ is particularly less
informative. At $k_\epsilon = 0.2$, we obtain interval widths around
$0.50$. As \cite{manski:07} emphasizes, this is the price for making
fewer assumptions. Even there, they typically cover only about
25\% of the interval $[-1, 1]$ of \emph{a priori} possibilities for the ACE.

\subsection{Influenza Study}
\label{sec:influenza}

Our empirical study concerns the effect of influenza vaccination on a
patient being later on hospitalized with chest problems. $X = 1$ means
the patient got a flu shot, $Y = 1$ indicates the patient was
hospitalized. A negative ACE therefore suggests a desirable
vaccine. The study was originally discussed by \cite{mcdonald:92}.
Shots were not randomized, but doctors were randomly assigned to
receive a reminder letter to encourage their patients to be
inoculated, an event recorded as binary variable {\it GRP}. This
suggests the standard IV model in Figure \ref{fig:summary}(d), with $W
= \text{\it GRP}$ and $U$ unobservable. That is, $W$ and $U$ are
independent because $W$ is randomized, and there are resonable
justifications to believe the lack of a direct effect of
letter randomization on patient hospitalization.
\cite{richardson:11} and \cite{hirano:00} provide further discussion.

From this randomization, it is possible to directly estimate the
ACE\footnote{Notice that while the ACE might be small, this does not
mean that in another scale, such as odd-ratios, the results do not
reveal an important effect. This depends on the domain.} of $W$ on
$Y$: $-0.01$. This is called {\it intention-to-treat} (ITT) analysis
\citep{rothman:08}, as it is based on the treatment assigned by
randomization and not on the variable of interest ($X$), which is not
randomized. While the ITT can be used for policy making, the ACE of $X$ on
$Y$ would be a more interesting result, as it reveals features of the
vaccine that are not dependent on the encouragement design. $X$ and
$Y$ can be confounded, as $X$ is not controlled. For instance, the
patient choice of going to be vaccinated might be caused by her
general health status, which will be a factor for hospitalization in
the future.

The data contains records of $2,681$ patients, with some demographic
indicators (age, sex and race) and some historical medical data (for
instance, whether the patient is diabetic).  A total of 9 covariates
is available. Using the bounds of \cite{balke:97} and observed
frequencies gives an interval of $[-0.23, 0.64]$ for the ACE. {\sc
WPP} could {\it not} validate {\it GRP} as a witness for any admissible set. 

Instead, when forbidding {\it GRP} to be included in an admissible set
(since the theory says {\it GRP} cannot be a common direct cause of
vaccination and hospitalization), {\sc WPP} selected as the
highest-scoring pair the witness {\it DM} (patient had history of
diabetes prior to vaccination) with admissible set composed of {\it
AGE} (dichotomized as ``60 or less years old,'' and ``above 60'') and
{\it SEX}. Choosing, as an illustration, $\epsilon_w = \epsilon_y =
\epsilon_x = 0.2$ and $\underline \beta = 0.9$, $\bar \beta = 1.1$, we
obtain the posterior expected interval $[-0.10, 0.17]$. This does {\it
not} mean the vaccine is more likely to be bad (positive ACE) than
good: the posterior distribution is over bounds, not over points,
being completely agnostic about the distribution within the
bounds. Notice that even though we allow for full dependence between
all of our variables, the bounds are stricter than in the standard IV
model due to the weakening of hidden confounder effects postulated by
observing conditional independences. It is also interesting that two
demographic variables ended up being chosen by Rule 1, instead of
other indicators of past diseases.

When allowing {\it GRP} to be included in an admissible set, the pair
({\it DM}, {\it AGE, SEX}) is now ranked second among all pairs that
satify Rule 1, with the first place being given by {\it RENAL} as the
witness (history of renal complications), with the admissible set
being {\it GRP, COPD} (history of pulmonary disease), and {\it
SEX}. In this case, the expected posterior interval was approximately
the same, $[-0.07, 0.16]$. It is worthwhile to mention that, even
though this pair scored highest by our criterion that measures the
posterior probability distribution of each premise of Rule 1, it is
clear that the fit of this model is not as good as the one with {\it
DM} as the witness, as measured by the much larger proportion of
rejected samples when generating the posterior distribution. This
suggests future work on how to rank such models.

\begin{figure}[t]
\begin{center}
\begin{tabular}{c}
\includegraphics[width=3in]{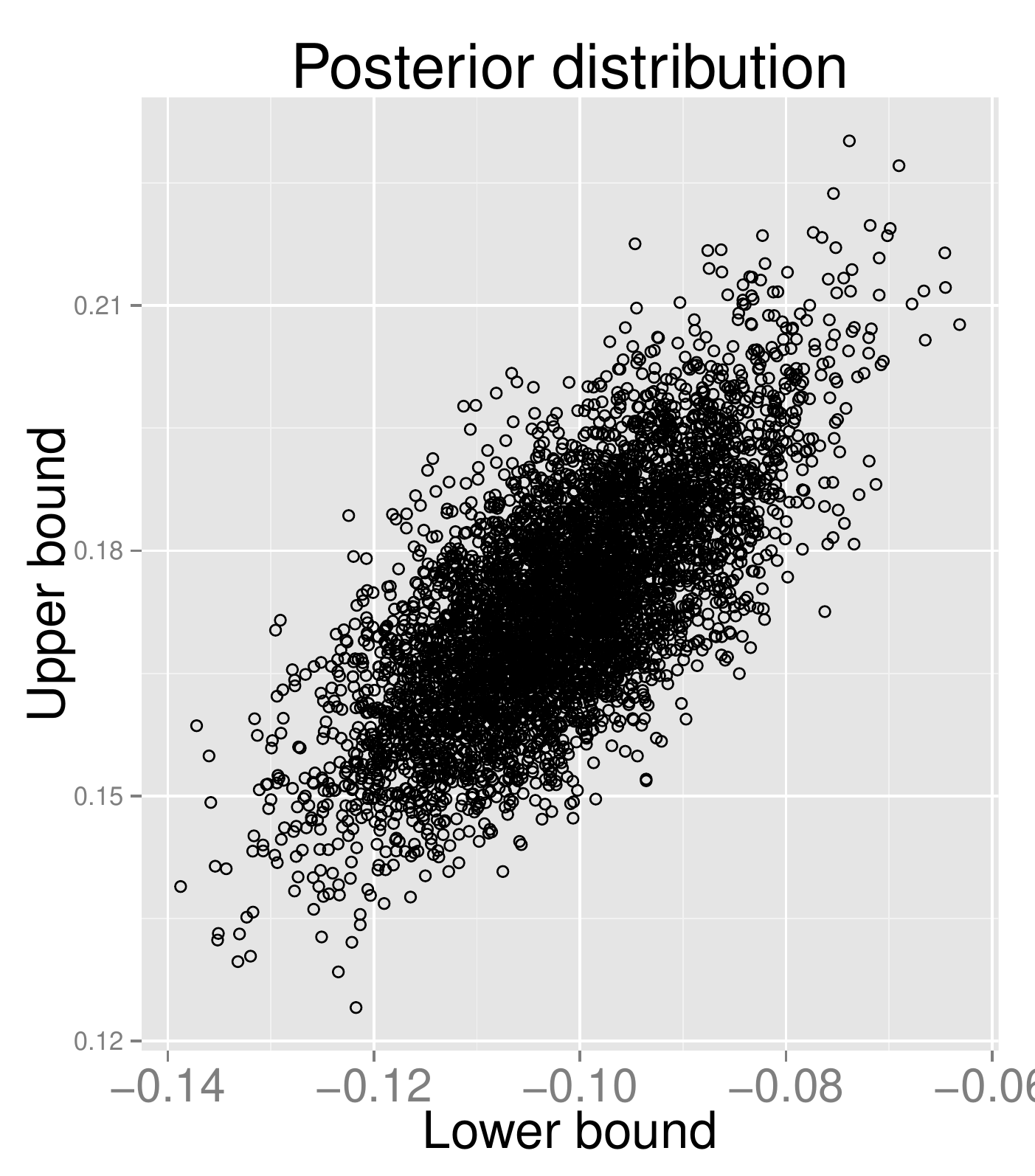}
\end{tabular}
\end{center}
\caption{Scatterplot of the joint posterior distribution of lower bounds and upper bounds,
Pearson correlation coefficient of 0.71.}
\label{fig:scatter_influenza}
\end{figure}

\begin{figure}[t]
\begin{center}
\begin{tabular}{cc}
\includegraphics[width=3in,height=3in]{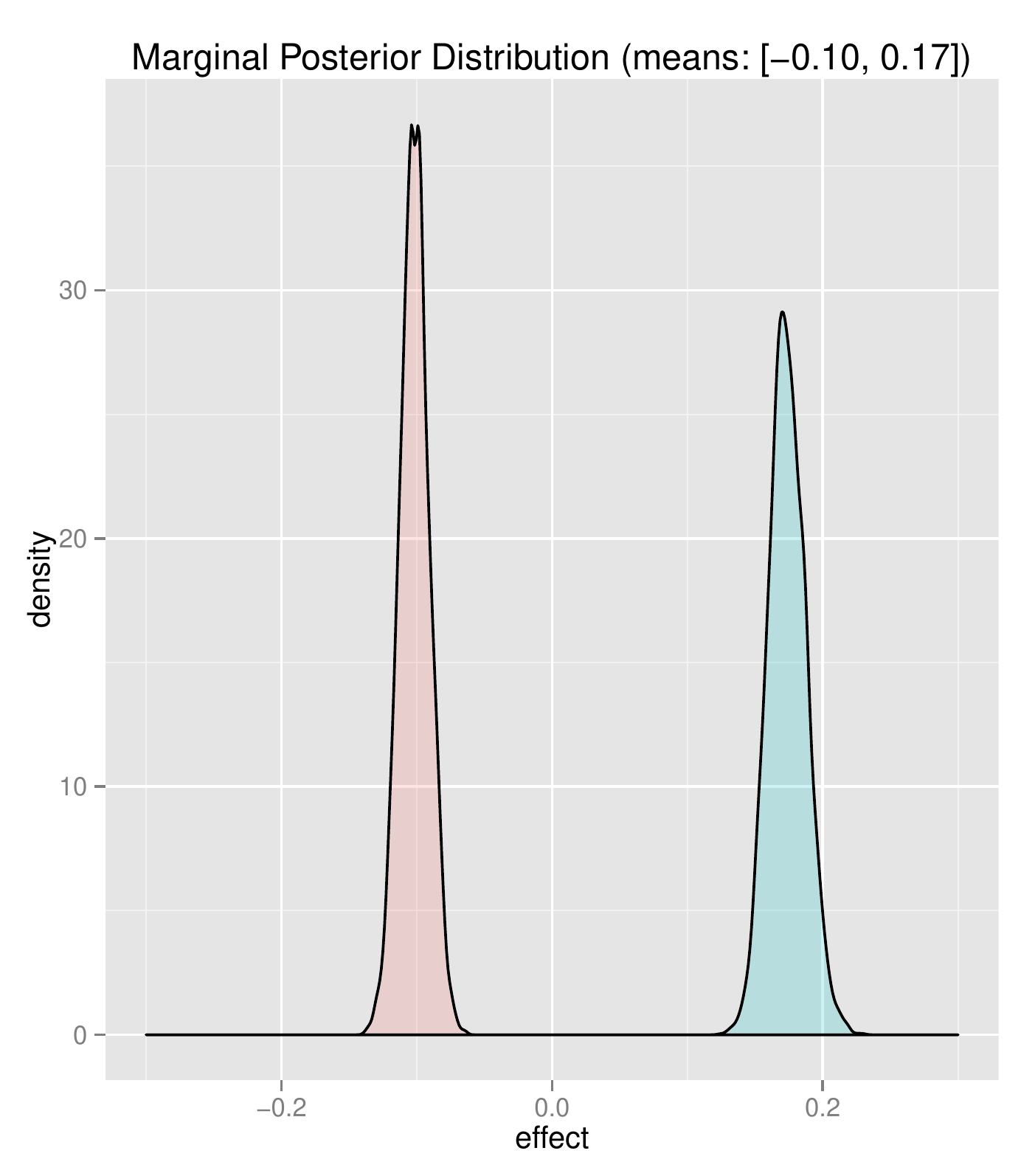} &
\includegraphics[width=3in,height=3in]{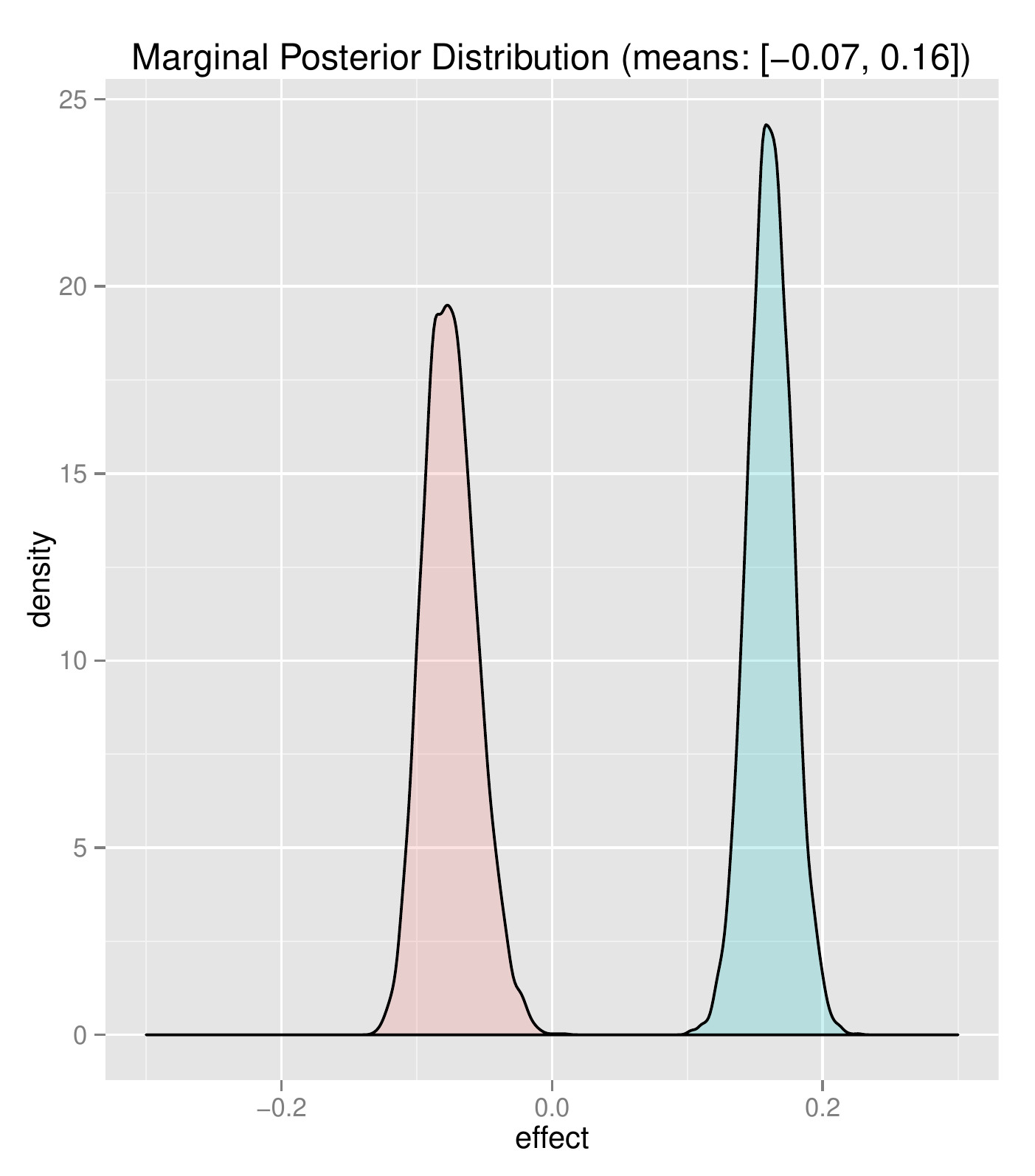} \\
(a) & (b)\\
\end{tabular}
\end{center}
\caption{In (a), the marginal densities for the lower bound (red) and
upper bound (blue) on the ACE, smoothed kernel density estimates based
on 5000 Monte Carlo samples. Bounds were derived using $DM$ as the
witness. In (b), a similar plot using $RENAL$ as the witness.}
\label{fig:marginal_influenza}
\end{figure}

In Figure \ref{fig:scatter_influenza} we show a scatter plot of the
posterior distribution over lower and upper bounds on the influenza
vaccination, where $DM$ is the witness. In Figure
\ref{fig:marginal_influenza}(a) and (b) we show kernel density
estimators based on the Monte Carlo samples for the cases where $DM$
and $RENAL$ are the witnesses, respectively. While the witnesses were
tested using the analytical bounds, the final set of samples shown
here were generated with the fully numerical optimization procedure,
which is quite expensive.

We also analyze how Algorithm \ref{algo:sel_xy} and its variants
for $\tau_w$ and $\tau_c$ can be used to select $\aleph =
\{\epsilon_w, \epsilon_x, \epsilon_y, \underline \beta, \bar \beta\}$.
The motivation is that this is a domain with overall weak dependencies
among variables. From one point of view, this is bad as instruments
will be weak and generate wide intervals (as suggested by Proposition
\ref{prop:siv}). From another perspective, this suggests that the effect
of hidden confounders may also be weak. 

\begin{figure}[t]
\begin{center}
\begin{tabular}{cc}
\includegraphics[width=3in,height=3in]{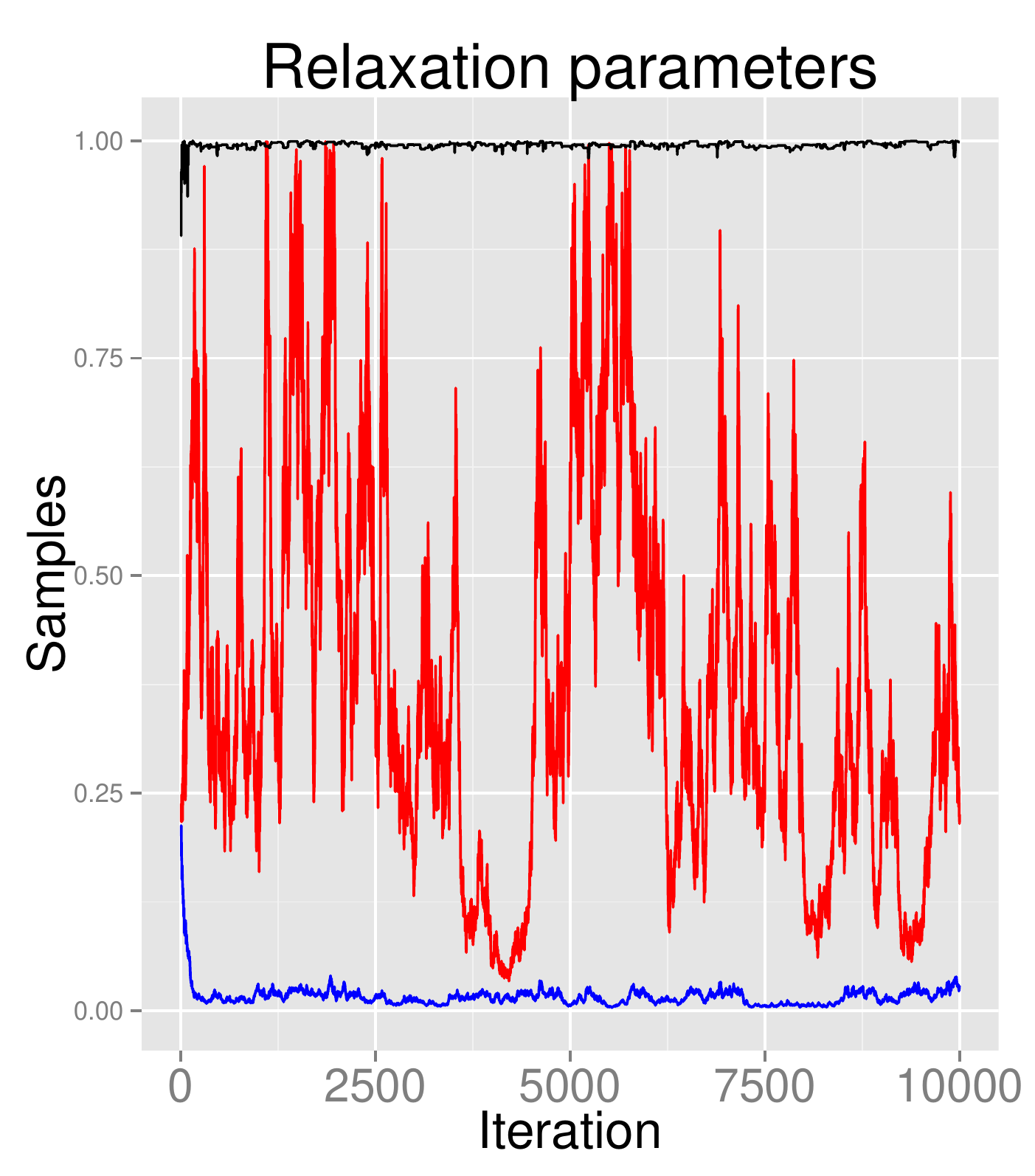} &
\includegraphics[width=3in,height=3in]{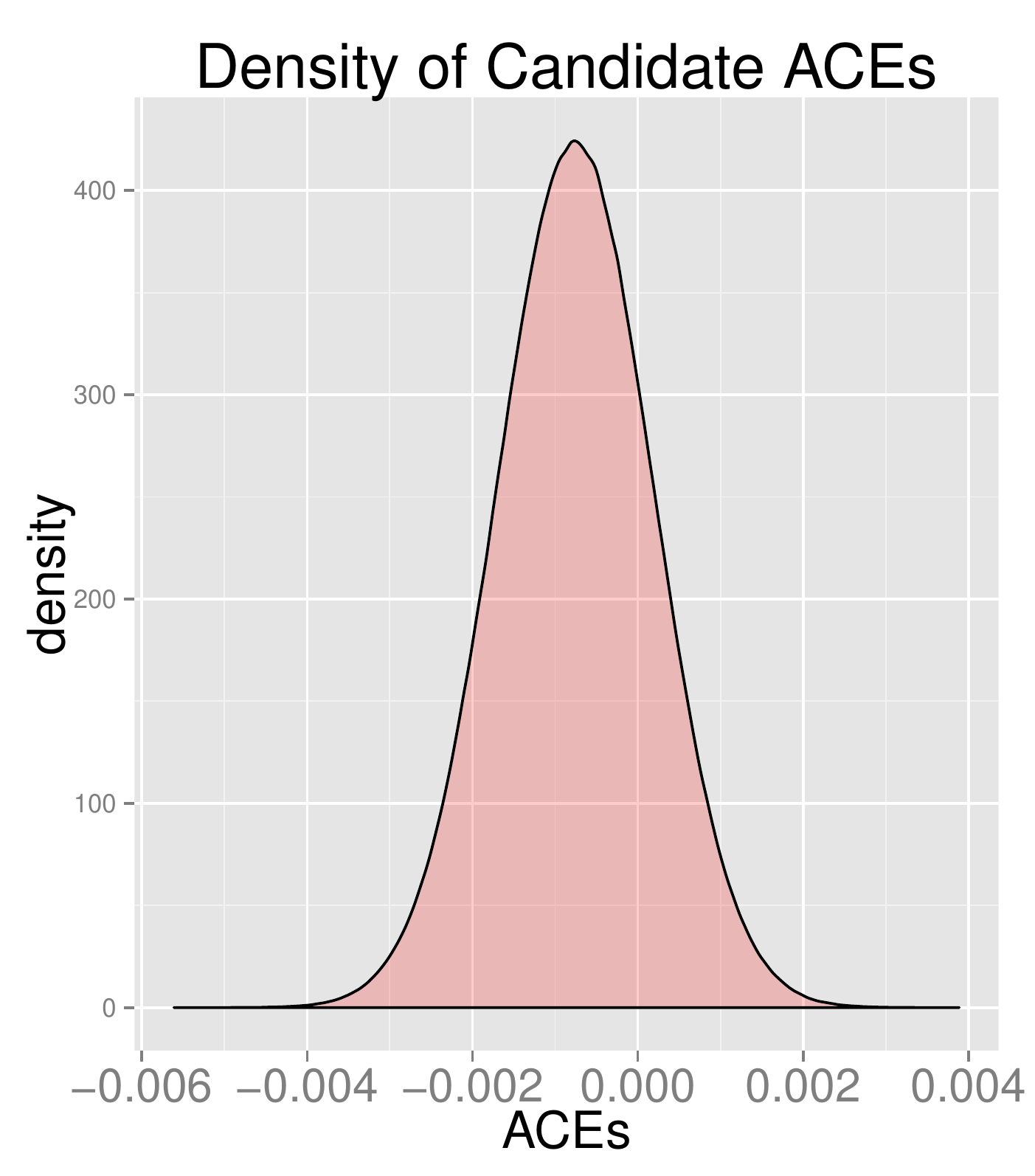} \\
(a) & (b)\\
\end{tabular}
\end{center}
\caption{In (a), MCMC plots for the relaxation parameters $\epsilon_w$ (red curve), $\epsilon_x = \epsilon_y$ (blue curve)
and $\underline \beta = 1 / \bar \beta$ (black curve) using the framework
of Section \ref{sec:selection}. The respective means are $0.38$,
$0.02$ and $0.99$.  In (b), a Gaussian fit for the chosen ACEs used to
generate the posterior over the relaxation parameters (mean $-0.0007$, standard deviation $0.0009$).}
\label{fig:aleph_analysis}
\end{figure}

Following the framework of Algorithm \ref{algo:sel_aleph} in Section
\ref{sec:selection}, we put independent uniform $[0, 1]$ priors on
the relaxation parameters $\epsilon_w, \epsilon_x = \epsilon_y$ and
$\underline \beta = 1 / \bar \beta$. 8 admissible sets provide
the reference set for the target set ({\it DM}, ({\it AGE}, {\it SEX})), where we
disallow the empty set and any admissible set containing {\it
GRP}. Reference set back-door adjusted ACEs are all very weak. Figure
\ref{fig:aleph_analysis} shows the posterior inference and a Gaussian
density estimate of the distribution of reference ACEs using the
empirical distribution to estimate each individual ACE.

The result is used to define strongly informative relaxation
parameters $0.56, 0.02$ and $0.99$, with a corresponding expected
posterior interval of $[0.01, 0.02]$, suggesting a deleterious effect of
the vaccination. The $95\%$ posterior credible interval for
the lower bound, however, also includes zero. While we do not claim by any
means that this procedure provides irrefutable ACE bounds for this
problem (such is the case for any observational study), this
illustrates that, even for a small number of covariates, there is an
opportunity to use reasonably broad priors and obtain informative
consequences on the values of $\aleph$ by a more conservative
exploitation of the faithfulness assumption of
\cite{sgs:00}.

\section{Conclusion}
\label{sec:conclusion}

Our model provides a novel compromise between point estimators given
by the faithfulness assumption and bounds based on instrumental
variables.  We believe such an approach should become a standard item
in the toolbox of anyone who needs to perform an observational
study\footnote{ {\sc R} code for all methods is available at
http://www.homepages.ucl.ac.uk/$\sim$ucgtrbd/wpp.}.

Unlike risky Bayesian approaches that put priors directly on the
parameters of the unidentifiable latent variable model $P(Y, X, W, U\,
|\, \mathbf Z)$, the constrained Dirichlet prior on the observed
distribution does not suffer from massive sensitivity to the choice of
hyperparameters. By focusing on bounds, {\sc WPP} keeps inference more
honest. While it is tempting to look for an alternative that will provide
a point estimate of the ACE, it is also important to have a method
that trades-off information for fewer assumptions. {\sc WPP} provides
a framework to express such assumptions.

As future work, we will look at a generalization of the procedure
beyond relaxations of chain structures $W \rightarrow X \rightarrow
Y$.  Much of the machinery here developed, including Entner et al.'s
Rules, can be adapted to the case where causal ordering is unknown:
starting from the algorithm of \cite{mani:06} to search for
``Y-structures,'' it is possible to generalize Rule 1 to setups where we
have an outcome variable $Y$ that needs to be controlled, but where there is no
covariate $X$ known not to be a cause of other covariates. Finally,
the techniques used to derive the symbolic bounds in Section
\ref{sec:scale-up} may prove useful in a more general context, and
complement other methods to find subsets of useful constraints such as
the graphical approach of
\cite{robin:12}.

\acks{We thank McDonald, Hiu and Tierney for their flu vaccine data.
Most of this work was done while RS was hosted by the Department of 
Statistics at the University of Oxford. }


\bibliography{rbas}

\appendix
\section*{Appendix A. Proofs}
\label{app:theorem}



In this Appendix, we prove the results mentioned in the main text.\\

\noindent {\bf Proof of Proposition \ref{prop:siv}} In the standard IV case, simple analytical
bounds are known for $P(Y = y\ |\ do(X = x))$ \citep{balke:97,dawid:03}:

\[
\begin{array}{rr}
\eta_0 \leq \min
\begin{cases}
  1 - \zeta_{00.0}\\
  1 - \zeta_{00.1}\\
  \zeta_{01.0} + \zeta_{10.0} + \zeta_{10.1} + \zeta_{11.1}\\
  \zeta_{10.0} + \zeta_{11.0} + \zeta_{01.1} + \zeta_{10.1}\\
\end{cases}
& 
\eta_0 \geq \max
\begin{cases}
  \zeta_{10.1}\\
  \zeta_{10.0}\\
  \zeta_{10.0} + \zeta_{11.0} - \zeta_{00.1} - \zeta_{11.1}\\
 -\zeta_{00.0} - \zeta_{11.0} + \zeta_{10.1} + \zeta_{11.1}\\
\end{cases}\\
&\\
\eta_1 \leq \min
\begin{cases}
  1 - \zeta_{01.1}\\
  1 - \zeta_{01.0}\\
  \zeta_{10.0} + \zeta_{11.0} + \zeta_{00.1} + \zeta_{11.1}\\
  \zeta_{00.0} + \zeta_{11.0} + \zeta_{10.1} + \zeta_{11.1}\\
\end{cases}
& 
\eta_1 \geq \max
\begin{cases}
  \zeta_{11.1}\\
  \zeta_{11.0}\\
 -\zeta_{01.0} - \zeta_{10.0} + \zeta_{10.1} + \zeta_{11.1}\\
  \zeta_{10.0} + \zeta_{11.0} - \zeta_{01.1} - \zeta_{10.1}\\
\end{cases}\\
\end{array}
\]

\noindent where $\eta_x \equiv P(Y = 1\ |\ do(X = x))$ and
$\zeta_{yx.w} \equiv P(Y = y, X = x\ |\ W = w)$. Define also
$\alpha_x \equiv P(Y = 1\ |\ X = x)$ and $\beta_w \equiv P(X = 1\ |\ W = w)$
so that 
\begin{equation}
\zeta_{yx.w} = \alpha_x^{I(y = 1)}(1 - \alpha_x)^{I(y = 0)}\beta_w^{I(x = 1)}(1 - \beta_w)^{I(x = 0)},
\label{eq:def_zeta_prop1}
\end{equation}
\noindent where $I(\cdot)$ is the indicator function
returning 1 or 0 depending on whether its argument is true or false, respectively.

Assume for now that $\beta_1 \geq \beta_0$, that is, $P(X = 1\ |\ W = 1) \geq P(X = 1\ | W = 0)$.
We will first show that $1 - \zeta_{00.0} \leq \min\{1 - \zeta_{00.1},
\zeta_{01.0} + \zeta_{10.0} + \zeta_{10.1} + \zeta_{11.1},  \zeta_{10.0} + \zeta_{11.0} + \zeta_{01.1} + \zeta_{10.1}\}$.

That $1 - \zeta_{00.0} \leq 1 - \zeta_{00.1}$ follows directly from the relationship (\ref{eq:def_zeta_prop1}) and
the assumptions $W \cind Y\ | X$ and $\beta_1 \geq \beta_0$: $(1 - \zeta_{00.0}) - (1 - \zeta_{00.1}) =
-(1 - \alpha_0)(1 - \beta_0) + (1 - \alpha_0)(1 - \beta_1) = (1 - \alpha_0)(\beta_0 - \beta_1) \leq 0$.

Now consider $(1 - \zeta_{00.0}) - (\zeta_{01.0} + \zeta_{10.0} + \zeta_{10.1} + \zeta_{11.1})$.
This is equal to 
\[
\begin{array}{rl}
 = & (1 - (1 - \alpha_0)(1 - \beta_0)) - ((1 - \alpha_1)\beta_0 + \alpha_0(1 - \beta_0) + \alpha_0(1 - \beta_1) + \alpha_1\beta_1)\\
 = & (\beta_0 + \alpha_0(1 - \beta_0)) - (\beta_0 - \alpha_1\beta_0 + \alpha_0(1 - \beta_0) + \alpha_0 - \alpha_0\beta_1 + \alpha_1\beta_1)\\
 = & \alpha_1(\beta_0 - \beta_1) - \alpha_0(1 - \beta_1) \leq 0
\end{array}
\]

Analogously, we can show that $1 - \zeta_{00.0} \leq \zeta_{10.0} + \zeta_{11.0} - \zeta_{01.1} - \zeta_{10.1}$. Tedious
but analogous manipulations lead to the overall conclusion

\[
\begin{array}{rr}
1 - \zeta_{00.0} = \min
\begin{cases}
  1 - \zeta_{00.0}\\
  1 - \zeta_{00.1}\\
  \zeta_{01.0} + \zeta_{10.0} + \zeta_{10.1} + \zeta_{11.1}\\
  \zeta_{10.0} + \zeta_{11.0} + \zeta_{01.1} + \zeta_{10.1}\\
\end{cases}
& 
\zeta_{10.0} = \max
\begin{cases}
  \zeta_{10.1}\\
  \zeta_{10.0}\\
  \zeta_{10.0} + \zeta_{11.0} - \zeta_{00.1} - \zeta_{11.1}\\
 -\zeta_{00.0} - \zeta_{11.0} + \zeta_{10.1} + \zeta_{11.1}\\
\end{cases}\\
&\\
1 - \zeta_{01.1} = \min
\begin{cases}
  1 - \zeta_{01.1}\\
  1 - \zeta_{01.0}\\
  \zeta_{10.0} + \zeta_{11.0} + \zeta_{00.1} + \zeta_{11.1}\\
  \zeta_{00.0} + \zeta_{11.0} + \zeta_{10.1} + \zeta_{11.1}\\
\end{cases}
& 
\zeta_{11.1} = \max
\begin{cases}
  \zeta_{11.1}\\
  \zeta_{11.0}\\
 -\zeta_{01.0} - \zeta_{10.0} + \zeta_{10.1} + \zeta_{11.1}\\
  \zeta_{10.0} + \zeta_{11.0} - \zeta_{01.1} - \zeta_{10.1}\\
\end{cases}\\
\end{array}
\]

The upper bound on the ACE $\eta_1 - \eta_0$ is obtained by
subtracting the lower bound on $\eta_0$ from the upper bound on
$\eta_1$. That is, $\eta_1 - \eta_0 \leq (1 - \zeta_{01.1}) -
\zeta_{10.0} = \mathcal U_{SIV}$.  Similarly, $\eta_1 - \eta_0 \geq
\zeta_{11.1} - (1 - \zeta_{00.0}) = \mathcal L_{SIV}$.  It follows
that $\mathcal U_{SIV} - \mathcal L_{SIV} = 1 - (P(X = 1\ |\ W = 1) -
P(X = 1\ |\ W = 0))$.

Finally, assuming $\beta_1 \leq \beta_0$ gives by symmetry the
interval width $1 - (P(X = 1\ |\ W = 0) - P(X = 1\ |\ W = 1))$,
implying the width in the general case is given by $1 - |P(X = 1\ |\ W
= 1) - P(X = 1\ |\ W = 0)|$. \hfill\BlackBox\\

Now we will prove the main theorems stated in Section
\ref{sec:scale-up}. To facilitate reading, we repeat here the
notation used in the description of the constraints with a few
additions, as well as the identities mapping different parameter
spaces and the corresponding assumptions exploited in the derivation.

We start with the basic notation,
\begin{center}
\[
\begin{array}{rcl}
\zeta_{yx.w}^\star & \equiv& P(Y = y, X = x \ |\ W = w, U)\\
\zeta_{yx.w} & \equiv& \sum_U P(Y = y, X = x \ |\ W = w, U)P(U\ |\ W = w)\\
           & = & P(Y = y, X = x\ |\ W = w)\\
\kappa_{yx.w} & \equiv& \sum_U P(Y = y, X = x \ |\ W = w, U)P(U)\\
\\
\eta_{xw}^\star & \equiv& P(Y = 1\ |\ X = x, W = w, U)\\
\eta_{xw} & \equiv& \sum_U P(Y = 1\ |\ X = x, W = w, U)P(U\ |\ W = w)\\
         & = & P(Y = 1\ |\ do(X = x), W = w)\\
\omega_{xw} & \equiv& \sum_U P(Y = 1\ |\ X = x, W = w, U)P(U)\\
\\
\delta_{w}^\star & \equiv& P(X = 1\ |\  W = w, U)\\
\delta_{w} & \equiv& \sum_U P(X = 1 \ |\  W = w, U)P(U\ |\ W) = P(X = 1\ |\  W = w)\\
          & = & \zeta_{11.w} + \zeta_{01.w}\\
\chi_{x.w} &\equiv& \sum_U P(X = x \ |\  W = w, U)P(U)\\
          &=& \kappa_{1x.w} + \kappa_{0x.w}\\
\\
\end{array}
\]
\end{center}

The explicit relationship between parameters describing the latent variable
model is:
\[
\begin{array}{rcl}
\zeta_{00.0}^\star & = & (1 - \eta_{00}^\star)(1 - \delta_0^\star)\\
\zeta_{01.0}^\star & = & (1 - \eta_{10}^\star)\delta_0^\star\\
\zeta_{10.0}^\star & = & \eta_{00}^\star(1 - \delta_0^\star)\\
\zeta_{11.0}^\star & = & \eta_{10}^\star\delta_0^\star\\
\zeta_{00.1}^\star & = & (1 - \eta_{01}^\star)(1 - \delta_1^\star)\\
\zeta_{01.1}^\star & = & (1 - \eta_{11}^\star)\delta_1^\star\\
\zeta_{10.1}^\star & = & \eta_{01}^\star(1 - \delta_1^\star)\\
\zeta_{11.1}^\star & = & \eta_{11}^\star\delta_1^\star\\
\end{array}
\]

All upper bound constants $U_{\cdot\cdot}^{\cdot U}$ are assumed to be
positive. For $L_{\cdot\cdot}^{\cdot U} = 0$, $c \geq 0$,
all ratios $c / L_{\cdot\cdot}^{\cdot U}$ are defined to
be positive infinite.

In what follows, we define ``the standard IV model'' as the one which
obeys exogeneity of $W$ and exclusion restriction -- that is, the
model following the directed acyclic graph $\{W \rightarrow X
\rightarrow Y, X \leftarrow U \rightarrow Y\}$. 
All variables are binary, and the goal is to bound
the average causal effect (ACE) of $X$ on $Y$ given a non-descendant
$W$ and a possible (set of) confounder(s) $U$ of $X$ and $Y$.\\

\noindent {\bf Proof of Theorem \ref{th:bound1}} Start with the relationship between $\eta_{xw}$ and its upper bound:
\[
\begin{array}{rcll}
\eta_{xw}^\star & \leq & U_{xw}^{YU} & \textrm{(Multiply both sides by $\delta_{x'.w}^\star$)}\\
\eta_{xw}^\star(1 - (1 - \delta_{x'.w}^\star)) & \leq & U_{xw}^{YU}\delta_{x'.w}^\star & \textrm{(Marginalize over $P(U)$)}\\
\omega_{xw} - \kappa_{1x.w} & \leq & U_{xw}^{YU}\chi_{x'.w} & \\
\omega_{xw} & \leq & \kappa_{1x.w}  + U_{xw}^{YU}(\kappa_{0x'.w} + \kappa_{1x'.w}) &               \\
\end{array}
\]
\noindent and an analogous series of steps gives $\omega_{xw} \geq
\kappa_{1x.w} + L_{xw}^{YU}(\kappa_{0x'.w} + \kappa_{1x'.w})$.  Notice
such bounds above will depend on how tight $\epsilon_y$ is.  As an
illustration of its implications, consider the derived identity
$\zeta_{0x.w}^\star = (1 - \eta_{xw}^\star)\delta_{x.w}^\star
\Rightarrow 1 - \eta_{xw}^\star =
\zeta_{0x.w}^\star/\delta_{x.w}^\star \Rightarrow 1 - \eta_{xw}^\star
\geq \zeta_{0x.w}^\star \Rightarrow \eta_{xw}^\star \leq 1 -
\zeta_{0x.w}^\star = \zeta_{0x.w}^\star + \zeta_{0x'.w}^\star +
\zeta_{1x'.w}^\star \Rightarrow \omega_{xw} \leq \kappa_{0x.w} +
\kappa_{0x'.w} + \kappa_{1x'.w}$. 

It follows from $U_{xw}^{YU} \leq 1$ that that the derived bound
$\omega_{xw} \leq \kappa_{1x.w} + U_{xw}^{YU}(\kappa_{0x'.w} +
\kappa_{1x'.w})$ is at least as tight as the one obtained via
$\eta_{xw}^\star \leq 1 - \zeta_{0x.w}^\star$. Notice also that the
standard IV bound $\eta_{xw} \leq 1 - \zeta_{0x.w}$
\citep{balke:97,dawid:03} is a special case for $\epsilon_y = 0$,
$\underline \beta = \bar \beta = 1$.

For the next bounds, consider
\[
\begin{array}{rcll}
\delta_{x.w}^\star & \leq & U_{xw}^{XU} &\\
\eta_{xw}^\star\delta_{x.w}^\star & \leq & U_{xw}^{XU}\eta_{xw}^\star & \textrm{(Marginalize over $P(U)$)}\\
\kappa_{1x.w} & \leq & U_{xw}^{XU}\omega_{xw} & \\
\omega_{xw} & \geq & \kappa_{1x.w} / U_{xw}^{XU} & \\ 
\end{array}
\]
\noindent where the bound $\omega_{xw} \leq \kappa_{1x.w} /
L_{xw}^{XU}$ can be obtained analogously.  The corresponding bound for
the standard IV model (with possible direct effect $W \rightarrow Y$)
is $\eta_{xw} \geq \zeta_{1x.w}$, obtained again by choosing
$\epsilon_x = 1$, $\underline \beta = \bar \beta = 1$. The
corresponding bound $\omega_{xw} \geq \kappa_{1x.w}$ is a looser bound
for $U_{xw}^{XU} < 1$. Notice that if $L_{xw}^{XU} = 0$, the upper bound is
defined as infinite.

Finally, the last bounds are similar to the initial ones, but as a function of
$\epsilon_x$ instead of $\epsilon_y$:
\[
\begin{array}{rcll}
\delta_{x.w}^\star & \leq & U_{xw}^{XU} &\\
(1 - \eta_{xw}^\star)\delta_{x.w}^\star & \leq & U_{xw}^{XU}(1 - \eta_{xw}^\star) & \textrm{(Marginalize over $P(U)$)}\\
\kappa_{0x.w} & \leq & U_{xw}^{XU}(1 - \omega_{xw}) & \\
\omega_{xw} & \leq & 1 - \kappa_{0x.w} / U_{xw}^{XU} & \\ 
\end{array}
\]
\noindent The lower bound
$\omega_{xw} \geq 1 - \kappa_{0x.w} / L_{xw}^{XU}$ is obtained
analogously, and implied to be minus infinite if $L_{xw}^{XU} = 0$. \hfill\BlackBox\\

\noindent {\bf Proof of Theorem \ref{th:bound2}}  We start with the following derivation,
\[
\begin{array}{rcll}
\eta_{xw'}^\star - \eta_{xw}^\star  & \leq & \epsilon_w &\\
\eta_{xw'}^\star\delta_{x.w'}^\star - \eta_{xw}^\star\delta_{x.w'}^\star  & \leq & \epsilon_w\delta_{x.w'}^\star & 
                           \textrm{(Use $-U_{xw'}^{XU} \leq -\delta_{x.w'}^\star$)}\\
\eta_{xw'}^\star\delta_{x.w'}^\star - \eta_{xw}^\star U_{xw'}^{XU}  & \leq & \epsilon_w\delta_{x.w'}^\star & 
                           \textrm{(Marginalize over $P(U)$)}\\
\kappa_{1x.w'} - \omega_{xw}U_{xw}^{XI} & \leq & \epsilon_w\chi_{x.w'} & \\
\omega_{xw} & \geq & (\kappa_{1x.w'} - \epsilon_w\chi_{x.w'}) / U_{xw'}^{XU} &  \\ 
\omega_{xw} & \geq & (\kappa_{1x.w'} - \epsilon_w(\kappa_{0x.w'} + \kappa_{1x.w'})) / U_{xw'}^{XU} &  \\ 
\end{array}
\]\\
\noindent Analogously, starting from $\eta_{xw'}^\star - \eta_{xw}^\star  \geq  \epsilon_w$,
we obtain $\omega_{xw} \leq (\kappa_{1x.w'} + \epsilon_w(\kappa_{0x.w'} + \kappa_{1x.w'})) / L_{xw'}^{XU}$.
Notice that for the special case $\epsilon_w$ and $U_{xw'}^{XU} = 1$, we obtain the corresponding
lower bound $\omega_{xw} \geq \kappa_{1x.w'}$ that relates $\omega$ and $\kappa$ across different
values of $W$.

The result corresponding to the upper bound $\eta_{xw} \leq 1 - \zeta_{0x.w'}$ can be obtained as follows:
\[
\begin{array}{rcll}
\eta_{xw'}^\star - \eta_{xw}^\star  & \geq & -\epsilon_w &\\
1 + \eta_{xw'}^\star - 1 - \eta_{xw}^\star  & \geq & -\epsilon_w &\\
(1 - \eta_{xw}^\star) - (1 - \eta_{xw'}^\star) & \geq & -\epsilon_w &\\
(1 - \eta_{xw}^\star)\delta_{x.w'}^\star - (1 - \eta_{xw'}^\star)\delta_{x.w'}^\star & \geq & -\epsilon_w\delta_{x.w'}^\star &\\
(1 - \eta_{xw}^\star)U_{xw'}^{XU} - (1 - \eta_{xw'}^\star)\delta_{x.w'}^\star & \geq & -\epsilon_w\delta_{x.w'}^\star & \textrm{(Marginalize over $P(U)$)}\\
(1 - \omega_{xw})U_{xw'}^{XU} - \kappa_{0x.w'} & \geq & -\epsilon_w\chi_{x.w'} &\\
\omega_{xw} & \leq & 1 - (\kappa_{0x.w'} - \epsilon_w(\kappa_{0x.w'} + \kappa_{1x.w'})) / U_{xw'}^{XU}\\
\end{array}
\]\\
\noindent with the corresponding lower bound (non-trivial for $L_{xw'}^{XU} > 0$) given by
$\omega_{xw}^\star \geq 1 - (\kappa_{0x.w'} + \epsilon_w(\kappa_{0x.w'} + \kappa_{1x.w'})) / L_{xw'}^{XU}$.

The final block of relationships can be derived as follows:
\[
\begin{array}{rcll}
\eta_{xw}^\star - \eta_{xw'}^\star & \leq & \epsilon_w &\\
\eta_{xw}^\star\delta_{x'.w}^\star - \eta_{xw'}^\star\delta_{x'.w}^\star  & \leq & \epsilon_w\delta_{x'.w}^\star &\\
\eta_{xw}^\star(1 - (1 - \delta_{x'.w}^\star)) - \eta_{xw'}^\star\delta_{x'.w}^\star  & \leq & \epsilon_w\delta_{x'.w}^\star &
                    \textrm{(Use $-U_{x'w}^{XU} \leq -\delta_{x'.w}^\star$)}\\
\eta_{xw}^\star - \eta_{xw}^\star(1 - \delta_{x'.w}^\star) - \eta_{xw'}^\star U_{x'.w}^{XU}  & \leq & \epsilon_w\delta_{x'.w}^\star &
                           \textrm{(Marginalize over $P(U)$)}\\
\omega_{xw} - \kappa_{1x.w} - \omega_{xw'}U_{x'w}^{XU}  & \leq & \epsilon_w\chi_{x'.w} &\\
\omega_{xw} - \omega_{xw'}U_{x'w}^{XU}  & \leq & \kappa_{1x.w} + \epsilon_w(\kappa_{0x'.w} + \kappa_{1x'.w}) & \\ 
\end{array}
\]
\noindent with the lower bound
$\omega_{xw} - \omega_{xw'}L_{x'w}^{XU} \geq \kappa_{1x.w} - \epsilon_w(\kappa_{0x'.w} + \kappa_{1x'.w})$
derived analogously. Moreover,
\[
\begin{array}{rcll}
\eta_{xw'}^\star - \eta_{xw}^\star & \leq & \epsilon_w &\\
(1 - \eta_{xw}^\star)\delta_{x'.w}^\star - (1 - \eta_{xw'}^\star)\delta_{x'.w}^\star & \leq & \epsilon_w\delta_{x'.w}^\star &\\
(1 - \eta_{xw}^\star)(1 - (1 - \delta_{x'.w}^\star)) - (1 - \eta_{xw'}^\star)U_{x'w}^{XU} & \leq & \epsilon_w\delta_{x'.w}^\star &\\
1 - \omega_{xw} - \kappa_{0x.w} - (1 - \omega_{xw'})U_{x'w}^{XU}& \leq & \epsilon_w\chi_{x'.w} &\\
\omega_{xw} - \omega_{xw'}U_{x'w}^{XU} & \geq & 1 - \kappa_{0x.w} - U_{x'w}^{XU} - \epsilon_w(\kappa_{0x'.w} + \kappa_{1x'.w}) &\\
\end{array}
\]
\noindent and the corresponding
$\omega_{xw} - \omega_{xw'}L_{x'w}^{XU} \leq 1 - \kappa_{0x.w} - L_{x'w}^{XU} + \epsilon_w(\kappa_{0x'.w} + \kappa_{1x'.w})$.
The last two relationships follow immediately from the definition of $\epsilon_w$. \hfill\BlackBox\\

Our constraints found so far collapse to some of the
constraints found in the standard IV models \citep{balke:97,dawid:03}
given $\epsilon_w = 0$, $\underline \beta = \bar \beta = 1$. Namely,
\[
\begin{array}{rcl}
\eta_{xw} & \leq & 1 - \zeta_{0x.w}\\
\eta_{xw} & \leq & 1 - \zeta_{0x.w'}\\
\eta_{xw} & \geq & \zeta_{1x.w}\\
\eta_{xw} & \geq & \zeta_{1x.w'}\\
\end{array}
\]

However, none of the constraints so far found counterparts in the following:
\[
\begin{array}{rcl}
\eta_{xw} & \leq & \zeta_{0x.w} + \zeta_{1x.w} + \zeta_{1x.w'} + \zeta_{1x'.w'}\\
\eta_{xw} & \leq & \zeta_{0x.w'} + \zeta_{1x.w'} + \zeta_{1x.w} + \zeta_{1x'.w}\\
\eta_{xw} & \geq & \zeta_{1x.w} + \zeta_{1x'.w} - \zeta_{0x.w'} - \zeta_{1x'.w'}\\
\eta_{xw} & \geq & \zeta_{1x.w'} + \zeta_{1x'.w'} - \zeta_{0x.w} - \zeta_{1x'.w}\\
\end{array}
\]

These constraints have the distinct property of being functions of
both $P(Y = x, X = x\ |\ W = w)$ and $P(Y = x, X = x\ |\ W = w')$,
simultaneously.  So far, we have only used the basic identities and
constraints, without attempting at deriving constraints
that are not a direct application of such identities.  In the
framework of \citep{dawid:03, ramsahai:12}, it is clear that general
linear combinations of functions of $\{\delta_{x.w}^\star\eta_{1x.w}^\star, \delta_{x.w}^\star,
\eta_{1x.w}^\star\}$ can generate constraints on observable quantities $\zeta_{yx.w}$
and causal quantities of interest, $\eta_{xw}$. We need to emcompass these possibilities
in a way we get a framework for generating symbolic constraints as a function of
$\{\epsilon_w, \epsilon_y, \epsilon_x, \underline \beta, \bar \beta\}$.

One of the difficulties on exploiting a black-box polytope package for
that is due to the structure of the process, which exploits the
constraints in Section \ref{sec:wpp} by first finding the extreme
points of the feasible region of $\{\delta_w^\star\}$,
$\{\eta_{xw}^\star\}$. If we use the constraints
\[
\begin{array}{c}
|\eta_{x1}^\star - \eta_{x0'}^\star| \leq \epsilon_w\\
0 \leq \eta_{xw}^\star \leq 1\\
\end{array}
\]
\noindent then assuming $0 < \epsilon_w < 1$, we always obtain the following six extreme points
\[
\begin{array}{c}
(0,  0)\\
(0,  \epsilon_w)\\
(\epsilon_w,  0)\\
(1 - \epsilon_w,  1)\\
(1,  1 - \epsilon_w)\\
(1,  1)\\
\end{array}
\]

In general, however, once we introduce constraints $L_{xw}^{YU} \leq
\eta_{xw}^\star \leq U_{xw}^{XU}$, the number of extreme points will
vary. Moreover, when multiplied with the extreme points of the space
$\delta_1^\star \times \delta_0^\star$, the resulting extreme points
of $\zeta_{yx.w}^\star$ might be included or excluded of the polytope
depending on the relationship among $\{\epsilon_w, \epsilon_x,
\epsilon_y\}$ and the observable $P(Y, X\ |\ W)$. Numerically, this is
not a problem (barring numerical instabilities, which do occur with a
nontrivial frequency). Algebraically, this makes the problem
considerably complicated\footnote{As a counterpart, imagine we defined
  a polytope through the matrix inequality $A\mathbf x \leq \mathbf
  b$. If we want to obtain its extreme point representation as an
  algebraic function of the entries of matrix $A$ and vector $\mathbf
  b$, this will be a complicated problem since we cannot assume we
  know the magnitudes and signs of the entries.}. Instead, in what
follows we will define a simpler framework that will not give tight
constraints, but will shed light on the relationship between
constraints, observable probabilities and the $\epsilon$
parameters. This will also be useful to scale up the full Witness
Protection Program, as discussed in the main paper.

\subsection*{Methodology for Cross-W Constraints}

Consider the standard IV model again, i.e., where $W$ is exogenous
with no direct effect on $Y$.  So far, we have not replicated anything
such as e.g. $\eta_1 \leq \zeta_{00.0} + \zeta_{11.0} + \zeta_{10.1} +
\zeta_{11.1}$. We call this a ``cross-W'' constraint, as it relates observables
under different values of $W \in \{0, 1\}$.
These are important when considering weakening the
effect $W \rightarrow Y$. The recipe for deriving them will be as
follows. Consider the template
\begin{equation}
\label{eq:basic_lin}
\delta_0^\star f_1(\eta_0^\star, \eta_1^\star) +
\delta_1^\star f_2(\eta_0^\star, \eta_1^\star) + 
f_3(\eta_0^\star, \eta_1^\star) \geq 0
\end{equation}
\noindent such that $f_i(\cdot, \cdot)$ are linear. Linearity is
imposed so that this function will correspond to a linear function of
$\{\zeta^\star, \eta^\star, \delta^\star\}$, of which expectations
will give observed probabilities or interventional probabilities.

We will require that evaluating this expression at each of the four
extreme points of the joint space $(\delta_0^\star, \delta_1^\star)
\in \{0, 1\}^2$ will translate into one of the basic constraints $1 -
\eta_i^\star \geq 0$ or $\eta_i^\star \geq 0$, $i \in \{0, 1\}$.  This
implies any combination of $\{\delta_0^\star, \delta_1^\star,
\eta_0^\star, \eta_1^\star\}$ will satisfy (\ref{eq:basic_lin}) (more on
that later).

Given a choice of basic constraint (say, $\eta_1^\star \geq 0$), and
setting $\delta_0^\star = \delta_1^\star = 0$, this immediately
identifies $f_3(\cdot, \cdot)$. We assign the constraint corresponding
to $\delta_0^\star = \delta_1^\star = 1$ with the ``complementary
constraint'' for $\eta_1$ (in this case, $\eta_1^\star \leq 1$). This leaves two
choices for assigning the remaining constraints. 

Why do we associate the $\delta_0^\star = \delta_1^\star = 1$ case
with the complementary constraint? Let us parameterize each function as
$f_i(\eta_0^\star, \eta_1^\star) \equiv a_i\eta_0^\star +
b_i\eta_1^\star + c_i$. Let $a_3 = q$, where either $q = 1$ (case
$\eta_0^\star \geq 0$) or $q = -1$ (case $1 - \eta_0^\star \geq 0$).
Without loss of generality, assume case $(\delta_0^\star = 1, \delta_1^\star
= 0)$ is associated with the complementary constraint where the coefficient of
$\eta_0^\star$ should be $-q$. For the other two cases, the coefficient of
$\eta_0^\star$ should be 0 by construction. We get the system
\[
\begin{array}{rcl}
a_3 & = & q\\
a_1 + a_3 & = & -q\\
a_2 + a_3 & = & 0\\
a_1 + a_2 + a_3 & = & 0\\
\end{array}
\]
This system has no solution. Assume instead $\delta_0^\star = 
\delta_1^\star = 1$ is associated with the complementary constraint
where the coefficient of $\eta_0^\star$ should be $-q$. The system now
is:
\[
\begin{array}{rcl}
a_3 & = & q\\
a_1 + a_3 & = & 0\\
a_2 + a_3 & = & 0\\
a_1 + a_2 + a_3 & = & -q\\
\end{array}
\]
This system always have the solution $a_1 = a_2 = -q$. We do have
freedom with $b_1, b_2, b_3$, which means we can choose to allocate
the remaining two cases in two different ways. 

\begin{lemma}
\label{lemma:aux}
Consider the constraints derived by the above procedure.
Then any choice of $(\delta_0^\star, \delta_1^\star,
\eta_0^\star, \eta_1^\star) \in [0, 1]^4$ will satisfy these constraints.
\end{lemma}

\noindent {\bf Proof}
Without loss of generality, let $f_3(\eta_0^\star,
\eta_1^\star) = q\eta_0^\star + (1 - q)/2$, $q \in \{-1, 1\}$. That is, $a_3 = q, b_3 = 0, c_3 = (1 - q) / 2$.
This implies $a_1 = a_2 = -q$ (as above). Associating
$(\delta_0^\star = 1, \delta_1^\star = 0)$ with $\eta_1^\star \geq 0$ gives
$\{b_1 = 1, c_1 = (q - 1) / 2\}$ and consequently associating
$(\delta_0^\star = 0, \delta_0^\star = 1)$ with $1 - \eta_1^\star \geq 0$
implies $\{b_2 = -1, c_2 = (1 + q) /2 \}$. 
Plugging this into the expression $\delta_0^\star f_1(\eta_0^\star, \eta_1^\star) +
\delta_1^\star f_2(\eta_0^\star, \eta_1^\star) + 
f_3(\eta_0^\star, \eta_1^\star)$ we get
\[
\begin{array}{rl}
 = & \delta_0^\star(-q\eta_0^\star + \eta_1^\star + (q - 1) / 2) + 
      \delta_1^\star(-q\eta_0^\star - \eta_1^\star + (1 + q) / 2) + 
      q\eta_0^\star + (1 - q) / 2\\
 = & \eta_0^\star(q - (\delta_0^\star + \delta_1^\star)q) +
     \eta_1^\star(\delta_0^\star - \delta_1^\star)+
     \delta_0^\star(q - 1) / 2 + \delta_1^\star(1 + q) / 2 + (1 - q)/2\\
 = & \eta_0^\star(q - (\delta_0^\star + \delta_1^\star)q) +
     \eta_1^\star(\delta_0^\star - \delta_1^\star) +
    (-q + (\delta_0^\star + \delta_1^\star)q) / 2 + (\delta_1^\star - \delta_0^\star + 1)/2\\
\\
 = & q((\delta_1^\star + \delta_0^\star) - 1)(1 - 2\eta_0^\star)/2 + 
   ((\delta_1^\star - \delta_0^\star)(1 - 2\eta_1^\star) + 1)/2\\
 =& (\delta_1^\star + \delta_0^\star - 1)s/2 + (\delta_1^\star - \delta_0^\star)t/2 + 1/2
\end{array}
\]
where $s = q(1 - 2\eta_0^\star) \in [-1,1]$ and $t = (1 -
2\eta_1^\star) \in [-1,1]$.  Then evaluating at the four extreme
points $s,t \in \{-1,+1\}$ we get $\delta_0, \delta_1, 1 - \delta_0,
1-\delta_1$, all of which are non-negative. \hfill\BlackBox\\

The procedure derives 8 bounds (4 cases that we get by associating $f_3$ with
either $\eta_x \geq 0$ or $1 - \eta_x \geq 0$. For each of these cases, 2
subcases what we get by assigning $(\delta_0^\star = 1, \delta_1^\star = 0)$ with
either $\eta_{x'} \geq 0$ or $1 - \eta_{x'} \geq 0$). 
Now, for an illustration of one case:\\

\noindent {\bf Deriving a constraint for the standard IV model, example: $f_3(\eta_0^\star, \eta_1^\star) \equiv \eta_0^\star \geq 0$} \\

\noindent Associate $\eta_1^\star \geq 0$ with assigment $(\delta_0^\star = 1, \delta_1^\star = 0)$
(implying we associate $\eta_1^\star \leq 1$ with assigment $(\delta_0^\star = 0, \delta_1^\star = 1)$
and $\eta_0^\star \leq 1$ with $(\delta_0^\star = 1, \delta_1^\star = 1)$).
This uniquely gives $f_1(\eta_0^\star, \eta_1^\star) = \eta_1^\star - \eta_0^\star$,
$f_2(\eta_0^\star, \eta_1^\star) = -\eta_1^\star -\eta_0^\star + 1$. The resulting expression is
\[
\delta_0^\star(\eta_1^\star - \eta_0^\star) + \delta_1^\star(-\eta_1^\star -\eta_0^\star + 1) + \eta_0^\star \geq 0
\]
\noindent from which we can verify that the assignment
$(\delta_0^\star = 1, \delta_1^\star = 1)$ gives $\eta_0^\star \leq
1$. Now, we need to take the expectation of the above with respect to
$U$ to obtain observables $\zeta$ and causal distributions $\eta$. However,
first we need some rearrangement so that we match $\eta_0^\star$ with
corresponding $(1 - \delta_w^\star)$ and so on.
\[
\begin{array}{rcl}
\eta_1^\star(\delta_0^\star - \delta_1^\star) + \eta_0^\star(1 - \delta_0^\star - \delta_1^\star) + \delta_1^\star &\geq& 0\\
\eta_1^\star(\delta_0^\star - \delta_1^\star) + \eta_0^\star((1 - \delta_0^\star) + (1 - \delta_1^\star) - 1) + \delta_1^\star &\geq& 0\\
\zeta_{11.0}^\star - \zeta_{11.1}^\star + \zeta_{10.0}^\star + \zeta_{10.1}^\star - \eta_0^\star + \zeta_{01.1}^\star + \zeta_{11.1}^\star &\geq& 0\\
\end{array}
\]
\noindent Taking expectations and rearranging it, we have
\[
\eta_0 \leq \zeta_{11.0} + \zeta_{10.0} + \zeta_{10.1} + \zeta_{01.1}
\]
\noindent rediscovering one of the IV bounds for $\eta_0$. Choosing to associate
$\eta_1^\star \geq 0$ with assigment $(\delta_0^\star = 0, \delta_1^\star = 1)$ will give instead
\[
\eta_0 \leq \zeta_{11.1} + \zeta_{10.1} + \zeta_{10.0} + \zeta_{01.0} 
\]

\noindent Basically the effect of one of the two choices within any case is to switch
$\zeta_{yx.w}$ with $\zeta_{yx.w'}$. \hfill\BlackBox\\

\subsection*{Deriving Cross-W Constraints}

What is left is a generalization of that under the condition $|\eta_{xw} -
\eta_{xw'}| \leq \epsilon_w$, $w \neq w'$, instead of $\eta_{xw} =
\eta_{xw'}$. In this situation, we exploit the constraint $\underline L \leq \eta_{xw}^\star \leq \bar U$ instead
of $0 \leq \eta_{xw}^\star \leq 1$ or $L_{xw}^{YU} \leq \eta_{xw}^\star \leq U_{xw}^{YU}$,
where $\underline L \equiv \min\{L_{xw}^{YU}\}, \bar U \equiv \max\{U_{xw}^{YU}\}$. 
Using $L_{xw}^{YU} \leq \eta_{xw}^\star \leq U_{xw}^{YU}$ complicates things considerably. 
Also, we will not derive here the analogue proof of Lemma 1 for the case where
$(\eta_0^\star, \eta_1^\star) \in [\underline L, \bar U]^2$, as it is analogous but with a
more complicated notation.\\

\noindent {\bf Proof of Theorem \ref{th:bound3}} We demonstrate this through two special cases.

\noindent \underline{General Model, Special Case 1:}
$f_3(\eta_{0w}^\star, \eta_{1w}^\star) \equiv \eta_{xw}^\star - \underline L \geq 0$ \\

\noindent There are two modifications. First, we perform the same associations as before, but with respect to
$\underline L \leq \eta_{xw}^\star \leq \bar U$ instead of
$0 \leq \eta_{x}^\star \leq 1$. Second, before we take expectations, we swap some of the $\eta_{xw}^\star$
with $\eta_{xw'}^\star$ up to some error $\epsilon_w$.

Following the same sequence as in the example for the IV model, we get the resulting
expression (where $x' \equiv \{0, 1\} \backslash x$):
\[
\delta_w^\star(\eta_{x'w}^\star - \eta_{xw}^\star) + \delta_{w'}^\star(-\eta_{x'w}^\star -\eta_{xw}^\star + \bar U + \underline L) 
+ \eta_{xw}^\star - \underline L \geq 0
\]
\noindent from which we can verify that the assignment
$(\delta_w^\star = 1, \delta_{w'}^\star = 1)$ gives $\bar U - \eta_{xw}^\star \geq
0$. Now, we need to take the expectation of the above with respect to
$U$ to obtain ``observables'' $\kappa$ and causal effects $\omega$. However,
the difficulty now is that terms $\eta_{xw}^\star\delta_{w'}^\star$ and 
$\eta_{xw'}^\star\delta_{w}^\star$ have no observable counterpart under expectation.
We get around this transforming $\eta_{xw'}^\star\delta_w^\star$ into 
$\eta_{xw}^\star\delta_w^\star$ (and $\eta_{xw}^\star\delta_{w'}^\star$ into 
$\eta_{xw'}^\star\delta_{w'}^\star$) by adding the corresponding correction
$-\eta_{xw}^\star \leq -\eta_{xw'}^\star + \epsilon_w$:

\[
\begin{array}{rcl}
\delta_w^\star(\eta_{x'w}^\star - \eta_{xw}^\star) + \delta_{w'}^\star(-\eta_{x'w}^\star -\eta_{xw}^\star + \bar U + \underline L) 
+ \eta_{xw}^\star - \underline L &\geq& 0\\
\delta_w^\star(\eta_{x'w}^\star - \eta_{xw}^\star) + \delta_{w'}^\star(-\eta_{x'w'}^\star + \epsilon_w -\eta_{xw'}^\star + \epsilon_w +
 \bar U + \underline L) + \eta_{xw}^\star - \underline L &\geq& 0\\
\eta_{x'w}^\star\delta_w^\star + 
\eta_{xw}^\star(1 - \delta_w^\star) -
\eta_{x'w'}\delta_{w'}^\star -
\eta_{xw'}\delta_{w'}^\star
+ \delta_{w'}^\star(\bar U + \underline L + 2\epsilon_w) - \underline L &\geq& 0
\end{array}
\]

\noindent Now, the case for $x = 1$ gives

\[
\begin{array}{rcl}
\eta_{0w}^\star\delta_w^\star + 
\eta_{1w}^\star(1 - \delta_w^\star) -
\eta_{0w'}\delta_{w'}^\star -
\eta_{1w'}\delta_{w'}^\star
+ \dots &\geq& 0\\
\eta_{0w}^\star(1 - (1 - \delta_w^\star)) + 
\eta_{1w}^\star(1 - \delta_w^\star) -
\eta_{0w'}^\star(1 - (1 - \delta_{w'}^\star)) -
\eta_{1w'}^\star\delta_{w'}^\star
+ \dots &\geq& 0
\end{array}
\]

\noindent Taking the expectations:
\begin{equation}
\omega_{0w} - \kappa_{10.w} +
\omega_{1w} - \kappa_{11.w} -
\omega_{0w'} + \kappa_{10.w'} -
\kappa_{11.w'} +
\chi_{w'}(\bar U + \underline L + 2\epsilon_w) - \underline L \geq 0
\end{equation}

\noindent Notice that for $\underline \beta = \bar \beta = 1$, $\underline L = 0$, $\bar U = 1$, $\epsilon_w = 0$, this 
implies $\eta_{xw} = \eta_{xw'}$ and this collapses to
\[
\eta_{0w} - \zeta_{10.w} +
\eta_{1w} - \zeta_{11.w} -
\eta_{0w'} + \zeta_{10.w'} -
\zeta_{11.w'} +
\delta_{w'} \geq 0
\]
\[
\eta_{1w} \geq \zeta_{10.w} + \zeta_{11.w} - \zeta_{10.w'} - \zeta_{01.w'} 
\]
\noindent which is one of the lower bounds one obtains under the standard IV model.

The case for $x = 0$ is analogous and gives
\begin{equation}
\omega_{0w'} \leq 
\kappa_{11.w} +
\kappa_{10.w} +
\kappa_{10.w'} -
\kappa_{11.w'} +
\chi_{w'}(\bar U + \underline L + 2\epsilon_w) - \underline L
\end{equation}
The next subcase is when we exchange the assignment of $(\delta_w^\star, \delta_{w'}^\star)$
to other constraints. We obtain the following inequality:
\[
\delta_{w'}^\star(\eta_{x'w}^\star - \eta_{xw}^\star) + \delta_{w}^\star(-\eta_{x'w}^\star -\eta_{xw}^\star + \bar U + \underline L) 
+ \eta_{xw}^\star - \underline L \geq 0
\]
\noindent which from an analogous sequence of steps leads to
\[
\begin{array}{rcl}
\delta_{w'}^\star(\eta_{x'w}^\star - \eta_{xw}^\star) + \delta_{w}^\star(-\eta_{x'w}^\star -\eta_{xw}^\star + \bar U + \underline L) 
+ \eta_{xw}^\star - \underline L &\geq& 0\\
\delta_{w'}^\star(\eta_{x'w'}^\star + \epsilon_w - \eta_{xw'}^\star + \epsilon_w) + 
\delta_{w}^\star(-\eta_{x'w}^\star -\eta_{xw}^\star +
 \bar U + \underline L) + \eta_{xw}^\star - \underline L &\geq& 0\\
\eta_{x'w'}^\star\delta_{w'}^\star -
\eta_{xw'}^\star\delta_{w'}^\star +
2\delta_{w'}^\star\epsilon_w -
\eta_{x'w}^\star\delta_{w}^\star +
\eta_{xw}^\star(1 - \delta_{w}^\star)
+ \delta_{w}^\star(\bar U + \underline L) - \underline L &\geq& 0
\end{array}
\]

For $x = 1$,

\[
\begin{array}{rcl}
\eta_{0w'}^\star\delta_{w'}^\star -
\eta_{1w'}^\star\delta_{w'}^\star +
\eta_{0w}^\star\delta_{w}^\star +
\eta_{1w}^\star(1 - \delta_{w}^\star)
+ \dots & \geq & 0\\
\eta_{0w'}^\star(1 - (1 - \delta_{w'}^\star)) -
\eta_{1w'}^\star\delta_{w'}^\star -
\eta_{0w}^\star(1 - (1 - \delta_{w}^\star)) +
\eta_{1w}^\star(1 - \delta_{w}^\star)
+ \dots & \geq & 0\\
\end{array}
\]

Taking expectations,
\begin{equation}
\omega_{0w'} - \kappa_{10.w'} -
\kappa_{11.w'} - \omega_{0w} +
\kappa_{10.w} +
\omega_{1w} - \kappa_{11.w}
+ 2\chi_{w'}\epsilon_w + \chi_{w}(\bar U + \underline L) - \underline L \geq 0
\end{equation}

For $x = 0$,

\[
\begin{array}{rcl}
\eta_{1w'}^\star\delta_{w'}^\star -
\eta_{0w'}^\star\delta_{w'}^\star +
\eta_{1w}^\star\delta_{w}^\star +
\eta_{0w}^\star(1 - \delta_{w}^\star)
+ \dots & \geq & 0\\
\eta_{1w'}^\star\delta_{w'}^\star   -
\eta_{0w'}^\star(1 - (1 - \delta_{w'}^\star)) -
\eta_{1w}^\star\delta_{w}^\star +
\eta_{0w}^\star(1 - \delta_{w}^\star)
+ \dots & \geq & 0\\
\kappa_{11.w'} - \omega_{0w'} + \kappa_{10.w'} -
\kappa_{11.w} + \kappa_{10.w}
+ 2\chi_{w'}\epsilon_w + \chi_{w}(\bar U + \underline L) - \underline L &\geq& 0
\end{array}
\]
\begin{equation}
\omega_{0w'} \leq
\kappa_{11.w'} + \kappa_{10.w'} -
\kappa_{11.w} + \kappa_{10.w}
+ 2\chi_{w'}\epsilon_w + \chi_{w}(\bar U + \underline L) - \underline L
\end{equation}\\

\noindent \underline{General Model, Special Case 2:} $f_3(\eta_{0w}^\star, \eta_{1w}^\star) \equiv \bar U - \eta_{xw}^\star \geq 0$ \\

\noindent Associate $\eta_{x'w}^\star \geq \underline L$ with assigment $(\delta_w^\star = 1, \delta_{w'}^\star = 0)$
(implying we associate $\eta_{x'w}^\star \leq \bar U$ with assigment $(\delta_w^\star = 0, \delta_{w'}^\star = 1)$
and $\eta_{xw}^\star \geq \underline L$ with $(\delta_w^\star = 1, \delta_{w'}^\star = 1)$).
The resulting expression is
\[
\delta_w^\star(\eta_{x'w}^\star + \eta_{xw}^\star - \bar U - \underline L) + 
\delta_{w'}^\star(-\eta_{x'w}^\star + \eta_{xw}^\star) + \bar U - \eta_{xw}^\star \geq 0
\]

Following the same line of reasoning as before, we get this for $x = 1$:
\begin{equation}
\omega_{0w} - \omega_{0w'} - \omega_{1w}
- \kappa_{10.w} + \kappa_{11.w} + \kappa_{10.w'} + \kappa_{11.w'}
- \chi_w(\bar U + \underline L) + 2\epsilon_w\chi_{w'} + \bar U \geq 0
\end{equation}

We get this for $x = 0$:
\begin{equation}
  \omega_{0w'} \geq
 -\kappa_{11.w} + \kappa_{10.w} + \kappa_{11.w'} + \kappa_{10.w'}
+ \chi_w(\bar U + \underline L) - 2\epsilon_w\chi_{w'} - \bar U 
\end{equation}

With the complementary assignment, we start with the relationship
\[
\delta_{w'}^\star(\eta_{x'w}^\star + \eta_{xw}^\star - \bar U - \underline L) + 
\delta_{w}^\star(-\eta_{x'w}^\star + \eta_{xw}^\star) + \bar U - \eta_{xw}^\star \geq 0
\]

For $x = 1$,
\begin{equation}
\omega_{0w'} - \omega_{0w} - \omega_{1w}
- \kappa_{10.w'} + \kappa_{11.w'} + \kappa_{10.w} + \kappa_{11.w}
+ \chi_{w'}(2\epsilon_w - \bar U - \underline L) + \bar U \geq 0
\end{equation}

For $x = 0$,
\begin{equation}
\omega_{0w'} \geq
- \kappa_{11.w'} + \kappa_{10.w'} + \kappa_{11.w} + \kappa_{10.w}
- \chi_{w'}(2\epsilon_w - \bar U - \underline L) - \bar U 
\end{equation}

Notice that the bounds obtained are asymmetric in $x$, i.e., we derive different
bounds for $\omega_{0w}$ and $\omega_{1w}$. Symmetry is readily obtained by the same
derivation where $\delta_{w}^\star$ is interpreted as $P(X = 0\ |\ W = w, U)$ and
$x$ is swapped with $x'$. \hfill\BlackBox\\

\end{document}